%% file: iclr2024_conference.tex
\documentclass{article} % For LaTeX2e
\usepackage{iclr2024_conference,times}

\input{math_commands.tex}

\usepackage{hyperref}
\usepackage{url}

\usepackage{listings}
\usepackage{xcolor}
\usepackage{times}
\usepackage{latexsym}
\usepackage{arydshln}

\usepackage[T1]{fontenc}
\usepackage[utf8]{inputenc}

\usepackage{microtype}

\usepackage{inconsolata}

\usepackage{times}
\usepackage{latexsym}
\usepackage{amsfonts}
\usepackage{amsmath}
\usepackage{amssymb}
\usepackage{multicol}
\usepackage{multirow}
\usepackage{xspace}
\usepackage{booktabs}
\usepackage{bbding}
\usepackage{array}
\usepackage{threeparttable}
\usepackage{tcolorbox}
\usepackage{tabularx}
\usepackage{enumitem}
\usepackage[linesnumbered,ruled,vlined]{algorithm2e}
\usepackage{xcolor,colortbl}
\usepackage{color}
\usepackage{setspace}
\usepackage{makecell}
\usepackage{listings}

\usepackage{xltabular}
\usepackage{tcolorbox}
\tcbuselibrary{breakable}

\newcommand{\modelname}{\textsc{AutoIF}\xspace}

\lstdefinestyle{python}{
    language=Python,
    basicstyle=\ttfamily\small,
    keywordstyle=\color{blue}\bfseries,
    commentstyle=\color{green},
    stringstyle=\color{red},
    numberstyle=\tiny\color{gray},
    showstringspaces=false,
    frame=single,
    breaklines=true,
    backgroundcolor=\color{lightgray!20}
}

\usepackage{booktabs} % 用于美化表格的横线

\usepackage[utf8]{inputenc} % allow utf-8 input
\usepackage[T1]{fontenc}    % use 8-bit T1 fonts
\usepackage{hyperref}       % hyperlinks
\usepackage{url}            % simple URL typesetting
\usepackage{booktabs}       % professional-quality tables
\usepackage{amsfonts}       % blackboard math symbols
\usepackage{nicefrac}       % compact symbols for 1/2, etc.
\usepackage{microtype}      % microtypography
\usepackage{xcolor}         % colors
\usepackage{graphicx}       % for table and figure
\usepackage{wrapfig}        % 用于环绕文本图表
\usepackage{subcaption}     % 用于环绕文本表格的标题
\usepackage{makecell}
\usepackage{soul}

\hypersetup{
    colorlinks=true,
    linkcolor=blue,
    citecolor=blue,
}

\definecolor{Ocean}{RGB}{129,194,250}

\definecolor{deepgreen}{RGB}{0, 70, 0}
\usepackage{cleveref}
\crefformat{section}{\S#2#1#3}
\crefformat{subsection}{\S#2#1#3}
\crefformat{subsubsection}{\S#2#1#3}
\crefrangeformat{section}{\S\S#3#1#4 to~#5#2#6}
\crefmultiformat{section}{\S\S#2#1#3}{ and~#2#1#3}{, #2#1#3}{ and~#2#1#3}
\crefmultiformat{subsection}{\S\S#2#1#3}{ and~#2#1#3}{, #2#1#3}{ and~#2#1#3}
\Crefformat{figure}{#2Fig.~#1#3}
\Crefmultiformat{figure}{Figs.~#2#1#3}{ and~#2#1#3}{, #2#1#3}{ and~#2#1#3}
\Crefformat{table}{#2Tab.~#1#3}
\Crefmultiformat{table}{Tabs.~#2#1#3}{ and~#2#1#3}{, #2#1#3}{ and~#2#1#3}
\Crefformat{appendix}{Appx.~\S#2#1#3}
\crefmultiformat{appendix}{Appx.~\S#2#1#3}{ and~#2#1#3}{, #2#1#3}{ and~#2#1#3}
\crefformat{algorithm}{Alg.~#2#1#3}
\Crefformat{equation}{Eq.~#2#1#3}

\newcommand{\xhdr}[1]{\vspace{0.3em}\noindent{{\bf #1.}}}

\title{Self-play with Execution Feedback: Improving Instruction-following Capabilities of Large Language Models}

\author{Guanting Dong\thanks{Work done during internship at Qwen team, Alibaba Inc.} \ , Keming Lu, Chengpeng Li$^*$, Tingyu Xia$^*$, Bowen Yu\thanks{Corresponding author}\\ \bf{Chang Zhou, Jingren Zhou}
\\
Qwen Team, Alibaba Inc. \\
\texttt{\{dongguanting.dgt,lukeming.lkm,lichengpeng.lcp\}@alibaba-inc.com}\\
\texttt{\{xiatingyu.xty, yubowen.ybw,ericzhou.zc,jingren.zhou\}@alibaba-inc.com}\\
}

\usepackage{graphicx} % 添加可能需要的图形包
\usepackage{multirow} % 添加multirow包
\usepackage{booktabs} % 用于\toprule、\midrule等命令
\usepackage{makecell} % 用于makecell命令
\usepackage{colortbl} % 用于颜色
\usepackage{xcolor}   % 用于颜色

\begin{document}

\maketitle

\begin{abstract}
One core capability of large language models~(LLMs) is to follow natural language instructions. 
However, the issue of automatically constructing high-quality training data to enhance the complex instruction-following abilities of LLMs without manual annotation remains unresolved. 
In this paper, we introduce \modelname, the first scalable and reliable method for automatically generating instruction-following training data. 
\modelname transforms the validation of instruction-following data quality into code verification, requiring LLMs to generate instructions, the corresponding code to check the correctness of the instruction responses, and unit test samples to verify the code's correctness.
Then, execution feedback-based rejection sampling can generate data for Supervised Fine-Tuning (SFT) and Reinforcement Learning from Human Feedback (RLHF) training.
\modelname achieves significant improvements across three training algorithms, SFT, Offline DPO, and Online DPO, when applied to the top open-source LLMs, Qwen2 and LLaMA3, in self-alignment and strong-to-weak distillation settings. Our code is publicly available at \url{https://github.com/QwenLM/AutoIF}.

\end{abstract}

\begin{figure*}[h]
    \centering
    % \vspace{-1em}
    \small    \includegraphics[width=0.98\linewidth]{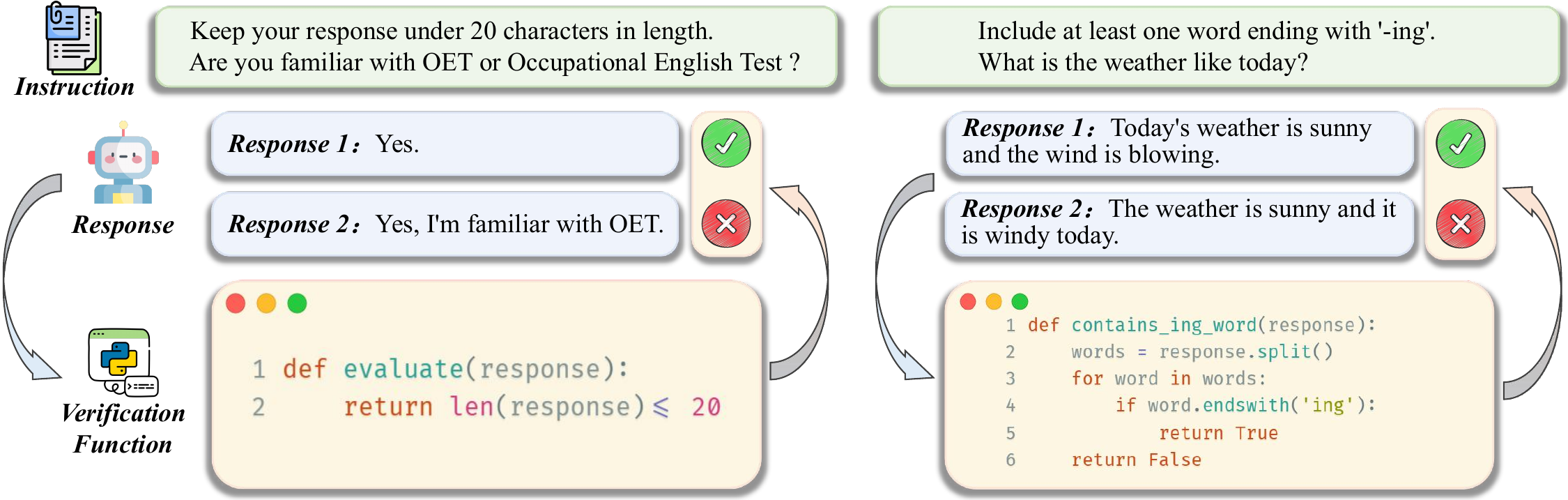}
    \caption{An example of the verification function automatically assesses the adherence of responses to the instruction's constraints.}
    \label{fig:intro}
    \vspace{-0.5em}
\end{figure*}

\section{Introduction}
\input{latex/intro}

% \begin{wrapfigure}{r}{0.48\textwidth} % 表示图形位于左侧,占据半个页面宽度
%   \centering % 图形居中
%   \vspace{-0.2cm}
%   \includegraphics[width=0.48\textwidth]{figures/case.pdf} % 替换为实际的图像文件
%   % \vspace{-0.4cm}
%   \caption{} % 图形标题
%   \label{}
% \end{wrapfigure}

\section{Related Works}
\input{latex/related}

\begin{figure*}[t]
    \centering
    \small    \includegraphics[width=0.98\linewidth]{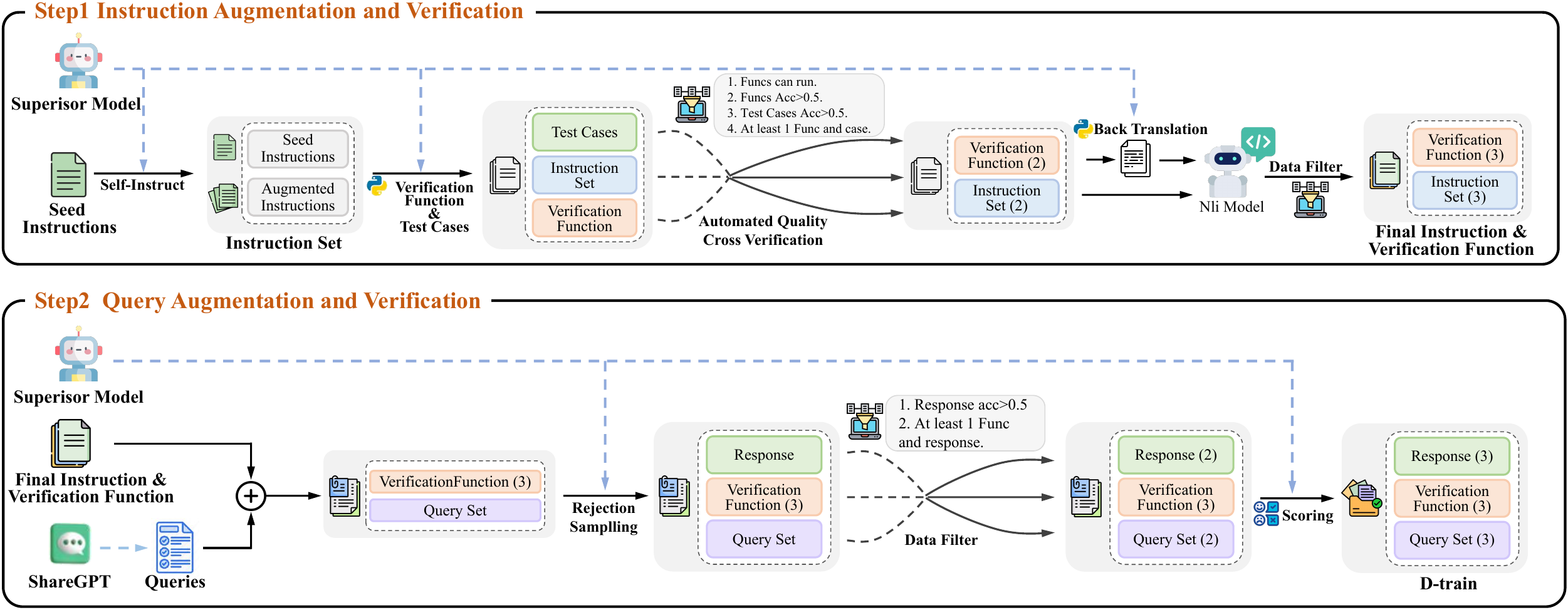}
    \caption{An Overview of \modelname: A Two-Stage Automated Instruction-Following Data Synthesis Method.}
    \label{main1}
    % \vspace{-1.5em}
\end{figure*}

\section{AutoIF}
\input{latex/method}

\section{Experiment}
\input{latex/experiments}

\section{Conclusion}
\input{latex/conclusion}

\bibliography{iclr2024_conference}
\bibliographystyle{iclr2024_conference}

\clearpage
\appendix
\section*{Appendix}
\label{sec:appendix}

\input{latex/appendix}

\end{document}

%% file: math_commands.tex
\usepackage{amsmath,amsfonts,bm}

\def\eqref#1{equation~\ref{#1}}
\def\1{\bm{1}}

\DeclareMathAlphabet{\mathsfit}{\encodingdefault}{\sfdefault}{m}{sl}
\SetMathAlphabet{\mathsfit}{bold}{\encodingdefault}{\sfdefault}{bx}{n}

%% file: latex/intro.tex
% Instruction-following capabilities of large language models~(LLMs) play a vital role in modern chatbots~\cite{openai2024gpt4,geminiteam2024gemini,bai2023qwen}, as advanced chatbots are expected to address user cases while precisely following diverse and complex instructions.

The instruction-following ability of large language models (LLMs) refers to their capacity to understand, interpret, and execute commands given to them in natural language~\citep{lou2023instruction, openai2024gpt4}. 
This ability is fundamental to contemporary LLMs as it enables them to leverage their underlying knowledge, interact intuitively with users~\citep{Ouyang2023rlhf}, adapt to various requirements~\citep{zhang2023instruction}, and perform complex tasks~\citep{sun2024conifer}. 
Misunderstandings in following instructions can lead to unintended outcomes, potentially resulting in severe consequences, particularly in critical scenarios~\citep{zhou2023instructionfollowing, chang2024survey}.

% \begin{figure}[t]
%     \centering
%     \includegraphics[width=\linewidth]{figures/case.pdf}

%     \caption{An example of the verification function automatically assesses the adherence of responses to the instruction's constraints.}
%     \label{fig:intro}
%     \vspace{-2em}
% \end{figure}

Although instruction following is crucial, scalable and reliable methods to enhance this capability of LLMs remain elusive. 
Current efforts in this field are divided into manual annotation~\citep{wei2021finetuned,zhou2023instructionfollowing,jiang2024followbench} and behavior imitation~\citep{xu2023wizardlm,zhao2024tree}.
Manual annotation involves annotators designing instructions and writing corresponding responses. 
However, due to human cognition's limitations, creating highly complex and diverse instructions is challenging, making the process difficult to scale. 
Furthermore, accurately executing complex instructions can sometimes be difficult for humans~\citep{sun2024conifer,cao2024towards}, requiring multiple rounds of rigorous and costly validation~\citep{wang2024human,wei2024long}. 
On the other hand, behavior imitation aims to distill responses from more advanced LLMs~\citep{taori2023stanford,peng2023instruction} like GPT-4. 
This approach limits models to the capabilities of the advanced LLMs from which they are distilled. 
Moreover, even advanced LLMs can make mistakes, and the reliability of the distilled data cannot be guaranteed~\citep{cui2023ultrafeedback}. 
Consequently, models trained with this data may have a propensity to not follow instructions accurately~\citep{zhou2024lima}.

In this paper, we introduce \modelname, the first scalable and reliable method for automatically generating instruction following training Data for Supervised Finetuning (SFT) or Reinforcement Learning from Human Feedback (RLHF)~\citep{Ouyang2023rlhf}. 
The core idea of \modelname is to use code to verify the correctness of following instructions.
Intuitively, if designed properly, a significant portion of instructions, such as “Keep your response under 20 characters in length” can be verified for correctness using code, as illustrated in \Cref{fig:intro}. 
Therefore, the key components of \modelname include (1) automatically generating instructions that can be verified by code, (2) automatically generating corresponding verification codes for these instructions, and (3) ensuring the reliability of the first two steps.
Specifically, we start by providing \modelname with a small set of hand-written seed instructions. 
Then, LLMs, not necessarily advanced ones, generate an augmented instruction set through self-instruct~\citep{wang2023selfinstruct}. 
Next, LLMs write verification codes and unit test cases for each instruction. 
Only the code that compiles correctly, passes the test cases, and back-translates to the original instruction is retained.
If an instruction does not have a corresponding code that can verify its correctness, it is discarded.
Finally, we employ LLMs to generate responses that either pass or fail the verification code using execution feedback-based rejection sampling~\citep{yuan2023scaling}. 
Responses that pass can be directly used for SFT, while pairs of passing and failing responses can be used to create chosen-rejected pairs for Direct Preference Optimization (DPO)~\citep{rafailov2023dpo} and other RLHF algorithms.
Moreover, once the instructions and verification code are determined, this process can be conducted on-policy, continually enhancing the instruction-following capabilities.

Through extensive experiments, we have demonstrated that \modelname achieves significant improvements under three training algorithms—SFT, Offline DPO, and Online DPO—when applied to the top open-source LLMs, Qwen2-72B and LLaMA3-70B, in both self-alignment and strong-to-weak distillation settings. 
In the IFEval benchmark, we achieved Loose Instruction (Acc.) rates of up to 88.0\% with Qwen2-72B and 90.4\% with LLaMA3-70B, marking the first instance of surpassing 90\% accuracy.
In the FollowBench benchmark, these models also showed significant improvements, with increases of over 5\% in the SSR metric (avg).
Additionally, they enabled Qwen2-7B and LLaMA3-8B to achieve average performance gains of over 4\% in both benchmarks.
Replacing Qwen2-72B and LLaMA3-70B with the more advanced GPT-4 resulted in further substantial improvements.
We will open-source the SFT and DPO datasets constructed using \modelname on Qwen2-72B, representing the first open-source complex instruction-following dataset at a scale of tens of thousands.

% In the IFEval and FollowBench instruction-following benchmarks, Qwen2-72B and LLaMA3-70B showes performance gains of up to [xx] and [xx] points, respectively. Additionally, they guided Qwen2-7B and LLaMA3-8B to improvements of [xx] points and [xx] points, respectively. 

%% file: latex/related.tex
\textbf{Instruction-following capabilities} are among the most essential features of  LLMs~\citep{openai2024gpt4,lou2023instruction}, which are expected to precisely follow a  broad and complex set of instructions. 
Consequently, recent research has concentrated on evaluating LLMs' instruction-following abilities in various contexts, such as verifiable~\citep{zhou2023instructionfollowing}, compositional~\citep{qin2024infobench}, format-related~\citep{xia2024fofo}, refuting~\citep{yan2024refutebench}, and fine-grained instructions~\citep{jiang2024followbench}.
However, a significant gap remains between open-source and proprietary closed-source LLMs. 
\citet{sun2024conifer} propose Conifer, which enhances the instruction-following capabilities of open-source LLMs through knowledge distillation from proprietary LLMs. 
\citet{wang2024codeclm} use LLMs to encode instruction metadata and augment diverse instructions from this metadata, employing proprietary LLMs for quality control. 
Both approaches, however, rely on proprietary LLMs for response distillation or judgment, which not only limits their potential but also subjects them to OpenAI's terms of use~\footnote{\url{https://openai.com/policies/terms-of-use}}.
In this work, we propose \modelname, a more scalable and reliable method to enhance the instruction-following capabilities of LLMs.
\modelname uses execution feedback from self-generated verification functions to provide supervision for instructions. 
This allows for effective self-alignment and strong-to-weak distillation on open-source models, thereby narrowing the performance gap with proprietary LLMs.

% Nevertheless, both works demand advanced LLMs for either response distillation or judgment, which still impairs their scalability.
% In this work, we propose \modelname, a much more scalable method to improve the instruction-following capabilities of LLMs.
% \modelname takes advantage of rejection sampling and utilizes execution feedback of self-generated verification functions to provide weak supervision for diverse instructions with an acceptable inference budget.

\textbf{Learning with Execution Feedback} is a widely-used technique in automated alignment for tool use and coding~\citep{cao2024scalable}.
These learning methods typically utilize execution feedback from tools such as code executors to provide supervision for specific tasks.
For instance, \citet{le2022coderl} employ feedback from unit tests via code compilers to enhance code synthesis capabilities through reinforcement learning.
Similarly, \citet{chen2023teaching} train LLMs to provide debugging suggestions as feedback to improve coding abilities.
Additionally, \citet{qiao2024making} introduce Reinforcement Learning with execution feedback to enhance LLMs using execution results from tools.
Building on this learning paradigm, we propose a novel scalable oversight method that enables LLMs to autonomously generate verification functions and unit tests for natural language instructions, thereby applying execution feedback to enhance their instruction-following capabilities.

%% file: latex/method.tex
We introduce \modelname, an automated, scalable, and reliable method designed to enhance the instruction-following capabilities of LLMs.
In this section, we outline the preliminaries (\Cref{sec:preliminaries}), detail the two core components of \modelname (\Cref{sec:ins}, \Cref{sec:query}), and discuss various training strategies that can be seamlessly integrated with \modelname (\Cref{sec:train}).

\subsection{Preliminaries}\label{sec:preliminaries}

\xhdr{Instruction-following Capabilities}
Following instructions is one of the most crucial skills in modern LLMs. 
These models are expected to provide precise responses to queries containing complex instructions, which can be either atomic or compositional. 
To evaluate the instruction-following capability of LLMs, we define a general instruction-following requirement as a specific task.
In this task, given an instruction $I = {\{i_{j}\}}_{j=1}^{N}$ with $N$ specific constraints  (e.g. “Please generate text in Shakespearean style, no more than 50 tokens” contains 2 constraints) and a specific query $x$, an LLM $\pi_{\theta}$ should generate precise response $y \sim \pi_{\theta}(y \mid x, I)$ adhering to the constraints.

\xhdr{Verifiable Instructions} 
The complexity and diversity of instructions necessitate manual construction and verification for reliable supervision.
This practical challenge motivates us to focus initially on instructions that can be automatically verified through programs and code executors, also known as verifiable instructions~\citep{zhou2023instructionfollowing}.
% Our experimental results indicate that training on instructions that can be automatically verified generally enhances the instruction-following capabilities of LLMs.
Specifically, for a given instruction $I$ and task-specific query $q$, there exists a verification function $f_{I}$ such that $f_{I}(y)$ returns true when the model's response $y$ correctly follows the instruction.
We demonstrate that supervision of such instructions can be self-generated through scalable oversight with LLMs and execution feedback. 
Extensive experiments in our work show that training on verifiable instructions significantly benefits the handling of other general instructions that are more complex but unverifiable with simple code snippets.

%Now assume we have a set of inputs $x \in D_{x}$ and candidate instructions $I \in D_{I}$, our goal is to generate instruction-following samples $(I,q, f_I, y) \in  D_{train}$, where $y \in p_{\theta}(y \mid q, I)$ and $f_I(y) = True$.

\xhdr{Method Overview} 
\modelname synthesizes high-quality instruction-following data through self-evolution, rejection sampling, and execution feedback. 
As illustrated in \Cref{main1}, \modelname integrates automated data augmentation with quality verification processes, including automatically generated verification functions and back-translation instructions.
This approach enables a two-stage automated data synthesis at both the instruction~(\Cref{sec:ins}) and query levels~(\Cref{sec:query}). 
Additionally, we introduce three training strategies~(\Cref{sec:train}) and explore two experimental settings~(\Cref{sec:setup}) to thoroughly evaluate the effectiveness and generalization of \modelname.

\subsection{Instruction Augmentation and Verification}
\label{sec:ins}

We first develop verifiable instructions along with corresponding evaluation functions, using rejection sampling informed by execution feedback.

\xhdr{Seed Instruction Construction}
We start by handwriting a set of seed instructions, denoted as $D_{seed}$,  ensuring that each instruction contains only a single atomic constraint (e.g., “Answer the words that begin with B”).
Detailed information on seed instructions is listed in \Cref{app:seed_instructions}.

% We begin with handwriting a set of seed instructions $D_{seed}$, controlling each instruction contains only an atomic constraint (e.g., “Answer the words that begin with B”). In addition, we manually check and make minor rephrases to ensure the diversity and uniqueness of each instruction. Detailed information on seed instructions is listed in \Cref{app:seed_instructions}.

\xhdr{Self-Instruct}
Self-Instruct~\citep{wang2023selfinstruct} is a straightforward and intuitive strategy for automated data augmentation that has garnered significant attention in the field of LLM reasoning~\citep{xu2023wizardlm,zhao2023preliminary}.
For each instruction in $D_{seed}$, we use an LLM to perform $K$ instruction rewrites, generating $D_{aug}$. We then combine the seed and augmented data sets to obtain an enhanced set of instructions, $D_{ins}=D_{seed} \cup D_{aug}$, and remove any duplicates.

\xhdr{Automated Quality Cross Verification}
Previous research has shown that relying solely on model-generated augmented instructions often leads to the inclusion of low-quality samples~\citep{mumuni2022data,xie2020unsupervised,zheng2024toward}. 
Inspired by a series of tool execution studies, we employ an LLM to generate verification functions and test cases for each instruction. We use feedback from executing Python programs to ensure quality control.
Given the instruction set $D_{ins}$, the LLM $M$ employs a rejection sampling~\citep{touvron2023LLaMA,yuan2023scaling} to generate $K$ verification functions $f_{I} = \{f_{i}\}_{j=1}^{K}$ and test cases $c_{I} = \{c_{i}\}_{j=1}^{K}$ for each instruction $I$, resulting in the set $\{I, f_{I}, c_{I}\} \in D_{ins}$. 
We then cross-validate the quality of the instructions using the verification functions and test cases, ensuring they meet the following criteria:
\begin{itemize}[leftmargin=1em]

\item The verification function $f \in f_{I}$ can be successfully compiled by the Python executor.
\item Each test case $c \in c_{I}$ achieves an accuracy rate greater than 0.5 across all verification functions.
\item Each verification function $f \in f_{I}$ achieves an accuracy rate greater than 0.5 across all test cases.
\item Each instruction includes at least one evaluation function and test case.
% \vspace{-0.5em}
\end{itemize}
By adhering to these four conditions, we obtain the quality-filtered instruction set $\{I^{(2)}, f_{I}^{(2)}\} \in D_{ins}^{(2)}$.

\xhdr{Back-translation Verification}
After the cross-validation stage, we obtained initially quality-verified verification functions and instructions. 
To further ensure the consistency between instructions and verification functions, we introduce back-translation.
For a given pair $\{I^{(2)}, f_{I}^{(2)}\} \in D_{ins}^{(2)}$, we use the LLM $M$ to back-translate the verification function $f \in f_{I}^{(2)}$ into instruction $I_f$.
We then treat $I$ as the \emph{premise} and the back-translated instruction $I_f$ as the \emph{hypothesis}. 
Using the NLI model, we identify the semantic relationship between the two instructions. 
The prediction can fall into one of three categories: \emph{entailment}, \emph{contradiction}, or \emph{neutral}:
\begin{equation}
p_{\theta}(\cdot \mid q, q_{\text{aug}})=\operatorname{softmax}\left({\operatorname{score}}_{\theta}(I, I_f)\right),
\end{equation}
where $\operatorname{score}_{\theta}: \mathbb{R}^{k \times \ell_{I}} \times \mathbb{R}^{k \times \ell_{I_f}} \rightarrow \mathbb{R}^{3}$ is a model dependent scoring function with parameters $\theta$. We filter out any instruction $I$ labeled as \emph{contradiction} 
%(approximately 20\%)
to ensure the intent consistency. 
Finally we obtain the set $\{I^{(3)}, f_{I}^{(3)}\} \in D_{ins}^{(3)}$

\begin{figure*}[t]
    \centering
    \small    \includegraphics[width=0.98\linewidth]{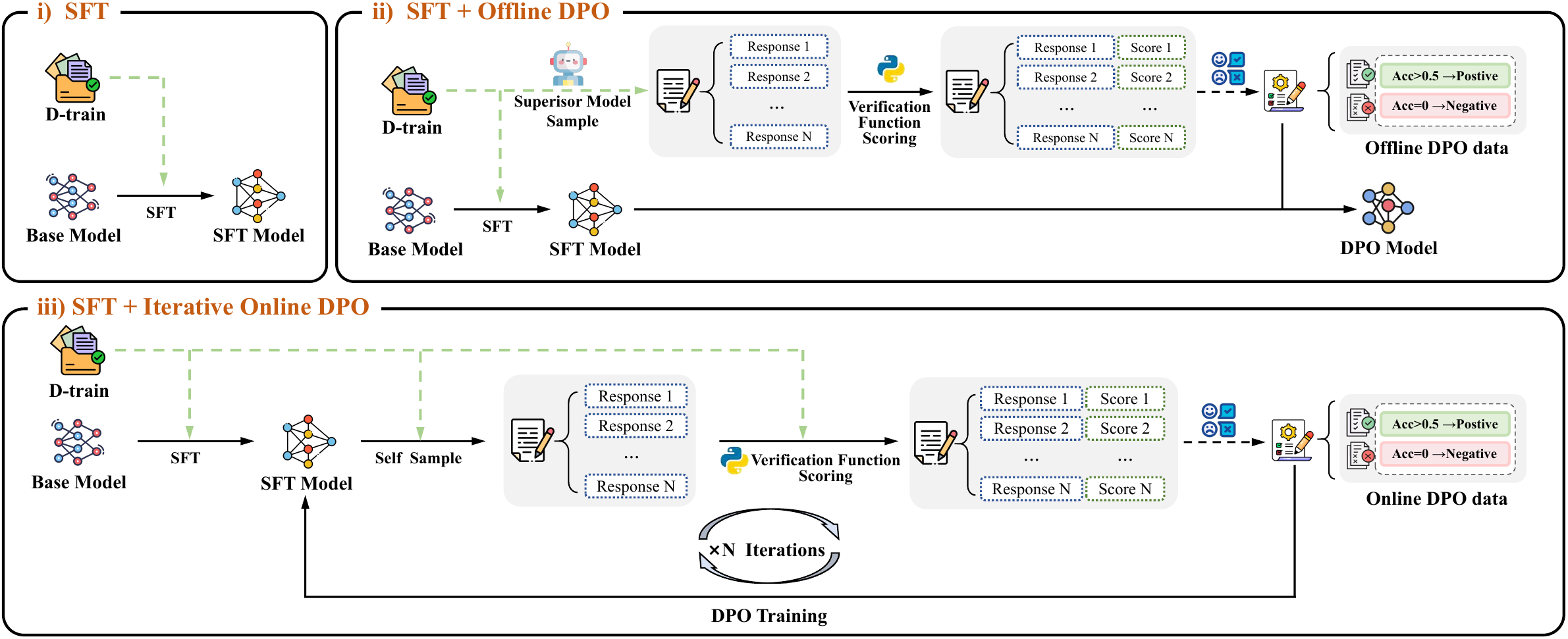}
        % \vspace{-0.5em}
    \caption{Different training strategies that can be adapted with synthetic dataset generated by  \modelname.}\vspace{-12pt}
    \label{main2}
    % \vspace{-1em}
\end{figure*}

\subsection{Query Augmentation and Verification}
\label{sec:query}

Once we have obtained verified instructions and verification functions, we utilize them to create training data comprising queries and responses.

\xhdr{Query Reforming and Augmentation}
In the real-world application of modern chatbots, instructions are typically employed to generate constrained responses to user queries. 
Therefore, creating high-quality instructions is merely the initial step toward achieving effective instruction-following capabilities. 
To acquire authentic queries, as shown in the bottom part of \Cref{main1}, we randomly selected $K$ user queries from ShareGPT~\citep{vicuna2023} for each instruction and concatenated them to construct the seed query dataset ${x, f_I^{(3)}} \in D_{q}$.  
To further enhance the diversity and complexity of the input $x$, we utilized the LLM to generate $K$ responses $y_x=\{y_i\}_{i=1}^K$, resulting in $\{x, f_I^{{3}}, y_x\} \in D_{q}$.

\xhdr{Instruction-following Verification}
Following the previous quality cross-verification process, we further employ verification functions to assess whether the augmented responses adhere to the constraints in input $x$. 
Similarly, we require each response in $D_{q}$ to meet the following conditions: 
\begin{itemize}[leftmargin=1em]

    \item Each response must achieve an accuracy rate greater than 0.5 across all verification functions.
    \item Each input must include at least one verification function and one response.

\end{itemize}
Based on these rules, we obtain the set $(x^{(2)},f_I^{(3)}, y^{(2)}) \in D_{q}^{(2)}$.

\xhdr{Query Quality Verification}
Additionally, we observe that concatenated instructions and queries often conflict. 
For instance, a high-quality response to the query ``help me write a news article'' is unlikely to comply with the instruction ``please limit your answer to two words''. 
Such high-level semantic inconsistencies are challenging for a simple NLI model to discern. 
Therefore, we employ the LLM $M$ to assign matching scores between the instruction and query in input $x^{(2)}$ and the corresponding responses $y^{(2)}$, on a scale from 1 to 10. 
We then filter out samples with a score lower than 8, constructing the final training set $D_{\text{train}} = \{x_i, y_i, f_{Ii}\}_{i=1}^N$.

\subsection{Training Strategies}
\label{sec:train}
\modelname offers multifaceted supervision for the instruction-following task, making it adaptable to various training strategies.
To thoroughly evaluate the effectiveness of \modelname, we propose the following training approaches:

\xhdr{Supervised Fine-tuning~(SFT)}
Given $(x_i, y_i) \in D_{\text{final}}$, we apply the standard Supervised Fine-tuning (SFT) objective on the base model $P$ with parameters $\theta$:
$
\label{eq:sft}
  \mathcal{L}(\boldsymbol{\theta})=\sum_{(x_i, y_i) \in \mathcal{D}_{\text{train}}} \log \mathbb{P_{\theta}}(y_i \mid x_i )
$
, where $x_i$ denotes the $i$-th input, consisting of a concatenated instruction and user query.

\xhdr{SFT + Offline DPO}
In the process of \modelname, multiple scales of quality filtering are utilized, naturally generating a substantial number of positive and negative sample pairs. 
This motivates us to obtain pairwise preference data $(x, y_w, y_l)$. Our preference data mining is divided into two parts:

\begin{itemize}[leftmargin=1em]

\item \textbf{Instruction Level:} During the automated quality cross-verification stage, we first extract positive samples $c_w$ from cases with an accuracy rate higher than 0.5 on all verification functions and negative samples $c_l$ from cases with an accuracy rate of 0. We then construct pairwise preference data for each instruction: $D_{\text{ins}}^{\text{pref}} \rightarrow (I, c_w, c_l)$.
\item \textbf{Query Level}: In the query quality verification process, we similarly extract positive samples $y_w$ from responses with an accuracy rate higher than 0.5 on all verification functions and negative samples $y_l$ from responses with an accuracy rate of 0. We then construct query preference data: $D_{\text{query}}^{\text{pref}} \rightarrow (x, y_w, y_l)$.

\end{itemize}

Finally, we merge the two parts of the data: 
$
D_{\text{pref}} = D_{\text{ins}}^{\text{pref}} \cup D_{\text{query}}^{\text{pref}}$. 
To further explore the potential of pairwise preference data $(x, y_w, y_l) \in D_{\text{pref}}$, we first perform vanilla SFT on the base model $\pi_\theta$ to obtain an SFT model $\pi_\theta^{\text{SFT}}$ as equation \ref{eq:sft}. Then, we apply Direct Preference Optimization (DPO)~\citep{rafailov2024direct} on our SFT model, which can be formulated as follows:

% \begin{small}
% \label{eq:dpo}
% \begin{gather*}
\begin{equation}
\mathcal{L}_{\text{DPO}}(\pi_\theta^{\text{SFT}};\pi_\text{ref}) = - \mathbb{E}_{(x, y_{w}, y_{l})\sim \mathcal{D}}[\text{log}\sigma (\beta \text{log}\frac{\pi_\theta^{\text{SFT}}(y_{w}|x)}{\pi_\text{ref}(y_{w}|x)}
- \beta \text{log}\frac{\pi_\theta^{\text{SFT}}(y_{l}|x)}{\pi_\text{ref}(y_{l}|x)})],
\end{equation}
% \end{gather*}
% \end{small}
% \end{equation}
where the reference model  $\pi_\text{ref}$ is set to $\pi_\theta^{\text{SFT}}$ initially and remains fixed throughout training. $\beta$ is a hyperparameter and $\sigma$ is the sigmoid function.
$\mathcal{L}_{\text{DPO}}$ aims to maximize the log probability of preferred $y_{w}$ relative to the dispreferred $y_{l}$.

\xhdr{SFT + Iterative Online DPO}
Online training enables real-time, iterative optimization of model weaknesses. 
It relies on high-quality, lightweight reward models to provide continuous supervision feedback. 
In the case of \modelname, verification functions serve as rigorous filtering standards, akin to reward models, delivering immediate feedback on model responses across training iterations. 
Following offline DPO, we conduct initial SFT on the base model $\pi_\theta$ to derive an SFT model $\pi_\theta^{\text{SFT}}$ with initial instruction-following capabilities.
As depicted in \Cref{main2}, we set the generation temperature to 0.8 and allow the SFT model to generate $K$ responses through self-sampling for each training sample, forming a response set $\{R_1, \ldots, R_k\}$. 
Then, we employ corresponding verification functions to assess $K$ responses, thereby constructing the online DPO dataset $D_{\text{online}}^{\text{pref}} = (x, y_w, y_l)$ based on average pass rates across all functions.
Finally, leveraging $D_{\textrm{online}}$, we sequentially perform DPO training on  $\pi_\theta^{\textrm{SFT}}$. Importantly, our iterative online optimization process progressively unlocks enhanced instruction-following capabilities.

%% file: latex/experiments.tex
\definecolor{deepgreen}{RGB}{0, 70, 0}
\definecolor{deepred}{RGB}{255, 0, 0}

\definecolor{backgreen}{RGB}{226, 240, 217}
\newcommand{\highg}{\cellcolor{backgreen}}

\subsection{Experimental Setup}\label{sec:setup}
\xhdr{Datasets \& Baselines} We conduct experiments using two LLMs from the Qwen2 series (Qwen2-7B and Qwen2-72B-Instruct) and two from the LLaMA3 series (LLaMA3-8B and LLaMA3-70B-Instruct).
Please note that the \modelname method proposed in this work has been employed in the open-source Qwen2-Instruct model. Thus, the version of Qwen2-Instruct we utilized represents an early iteration during internal development, rather than the final open-source model.
The training datasets are respectively generated from Qwen2-72B-Instruct and LLaMA3-70B-Instruct, with detailed statistics provided in \Cref{tab:efficiency}.
We demonstrate the effectiveness of \modelname by evaluating the instruction-following capabilities of models fine-tuned with self-generated datasets using \modelname. 
Additionally, we include strong open and closed-source LLM baselines such as Mixtral-8x22B and GPT-4. 
For more details, refer to \Cref{app:hyperparameters}.

\xhdr{Settings}
In our experiments, we mainly explore two experimental setups:

(1) \textbf{Strong-to-Weak Distillation} involves aligning a less powerful model with a stronger, well-aligned model by mimicking its generated responses. In \modelname, we can utilize a strong model such as Qwen2-72B-Instruct for data synthesis. Subsequently, we train a less powerful model like Qwen2-7B-Instruct using this synthesized data to achieve strong-to-weak alignment.

(2) \textbf{Self-Alignment}: Following several self-alignment works ~\citep{chen2024selfplay,yuan2024selfrewarding}, we utilize the LLM to perform the \modelname process for synthesizing data, and then train the same model using this synthesized data.

\xhdr{Evaluation}
We evaluate our methods using two instruction-following benchmarks: IFEval~\citep{zhou2023instructionfollowing} and FollowBench~\citep{jiang2024followbench}. 
IFEval comprises 25 types of verifiable instructions across about 500 prompts. 
While IFEval also focuses on verifiable instructions, extensive n-gram probing confirms no overlap between the IFEval test set and our training sets, thus eliminating any contamination concerns.
We report strict and loose accuracy metrics at both prompt and instruction levels for IFEval.
FollowBench is a fine-grained constraint-following benchmark with five levels of difficulty. 
It contains diverse and open-ended instructions requiring evaluation by strong LLMs, such as GPT-4, which can fully examine the generalization of \modelname to more general instructions not verifiable by simple code executions.
At the same time, we also evaluate our models on C-Eval~~\citep{huang2023ceval}, MMLU~~\citep{hendrycks2021measuring}, GSM8k~~\citep{cobbe2021training}, and HumanEval~~\citep{chen2021codex} to obtain a comprehensive assessment of capabilities.
% We evaluate our methods on two instruction-following benchmarks, IFEval~~\citep{zhou2023instructionfollowing} and FollowBench~~\citep{jiang2024followbench}.
% IFEval contains 25 types of verifiable instructions on around 500 prompts.
% It is worth noticing that IFEval is considered as an out-of-distribution evaluation for \modelname even though it also focuses on the evaluation of verifiable instructions, as IFEval contains compositional instructions, and extensive probing shows there is no contamination between the IFEval test set and any of our training set.
% We report both strict and loose accuracy on prompt and instruction levels for IFEval.
% FollowBench is a fine-grained constraint following benchmark with five levels of difficulties.
% FollowBench contains more diverse and open-ended instructions that require evaluation by strong LLMs, such as GPT-4-Turbo, which can fully examine the generalization of \modelname to more general instructions that are not verifiable by simple code executions.
% We report the average accuracy across five levels and each level for a more detailed analysis.

% "ifeval/strict_prompt_acc": 0.7855822550831792,
%         "ifeval/loose_prompt_acc": 0.8465804066543438,
%         "ifeval/strict_instruction_acc": 0.8525179856115108,
%         "ifeval/loose_instruction_acc": 0.89568345323741,

%----------------------------ablation---------------------------

%表x展示了我们我们baselines与不同setup下我们模型的效果，我们有以下几点观察
%

%----------------------------------main------------------------------------
\begin{table}[t]
\centering
\tiny
\caption{The main results on two instruction-following and four general benchmarks.
Pr. and Ins. stand for prompt and instruction levels, respectively. 
S and L represent strict and loose metrics for IFEval. The subscript indicates the increase in metrics compared to the corresponding backbone model. The highest accuracy for each setup is highlighted in  \colorbox{backgreen}{green}. 
Results marked with $^\dagger$ are directly sourced from the original benchmarks.}
\label{tab:main}
\setlength{\tabcolsep}{0.2mm}{%
\begin{tabular}{lcccccccccc|ccccc}
\toprule

\multirow{2}{*}{\textbf{Model}}   & \multicolumn{4}{c}{\textbf{IFEval}}       & \multicolumn{6}{c}{\textbf{FollowBench (SSR)}} &\multirow{2}{*}{\textbf{C-Eval}} &\multirow{2}{*}{\textbf{MMLU}} &\multirow{2}{*}{\textbf{GSM8k}}&\multirow{2}{*}{\textbf{HumanEval}}

\\
\cmidrule(lr){2-5} \cmidrule(lr){6-11}
      & \multicolumn{1}{c}{Pr~(S)}
      & \multicolumn{1}{c}{Pr.~(L)}
      & \multicolumn{1}{c}{Ins.~(S)}
      & \multicolumn{1}{c}{Ins.~(L)}
      & \multicolumn{1}{c}{Level 1}
      & \multicolumn{1}{c}{Level 2}
      & \multicolumn{1}{c}{Level 3}
      & \multicolumn{1}{c}{Level 4}
      & \multicolumn{1}{c}{Level 5}
      & \multicolumn{1}{c}{Avg} \\
\midrule

% 56.13%	52.67%	50.84%	45.19%	47.86%	50.54%

\multicolumn{11}{l}{\textit{Baselines (< 10B)}} \\
Qwen2-7B & 37.7 & 43.6 & 49.4 & 53.4 & 55.6 & 53.5 & 53.7 & 49.9 & 48.6 & 52.3    & 74.4 &	64.4 &	71.1  &	58.1\\  
Qwen2-7B(ShareGPT)  & 30.9 & 33.5 & 42.4 & 45.2 &56.1  &52.7  &50.8  &45.2  &47.9  &50.5 &70.2 & 59.8 & 59.4 & 52.4 \\
LLaMA3-8B & 24.6 & 26.1 & 38.1 & 39.7 &  10.0 & 10.3 & 10.5 & 14.3 & 12.7  & 11.6 &24.2 &38.8 & 4.5& 0.6 \\

LLaMA3-8B(ShareGPT)  & 23.7 & 26.4 & 33.8 & 37.1& 44.0  & 40.0 &39.6  &33.3  &33.6  &38.1 &35.2 & 44.6 & 20.5 & 38.1  \\

Mistral-7B & 23.3 & 24.6 & 38.4 & 39.6 & 40.1 & 39.7 & 37.9 & 35.7 & 36.7 & 38.0& 38.2 & 47.6 & 20.5 & 38.4  \\

% \cdashline{1-15}
\midrule

\multicolumn{11}{l}{\textit{Baselines (> 10B)}} \\
Qwen2-72B-Instruct & 77.1 & 80.4 & 84.4 & 86.9 & 70.2& 66.6 &63.5& 58.1 &56.3  &62.9 & 83.8 &80.8 & 87.9 & 73.8 \\
LLaMA3-70B-Instruct & 77.8 & 83.8 & 84.2 & 88.8 & 60.7 & 60.5 & 61.1 & 61.7 & 60.3 & 60.9 &60.2 &80.5 &92.6  &78.7 \\
Mixtral-8x22B & 41.8 & 47.3 & 55.2 & 60.0 & 63.9 & 60.0 & 58.2 & 56.2 & 55.3 & 58.7 & - & - & - & - \\
GPT-4$^\dagger$ & 76.9 & 79.3 & 83.6 & 85.4 & 84.7 & 77.6 & 76.2 & 77.9 & 73.3 & 77.9 & - & - & - & - \\
GPT-3.5 Turbo$^\dagger$ & - & - & - & - & 80.3 & 71.2 & 74.2 & 69.6 & 67.1 & 72.5& - & - & - & -  \\
\midrule
% ${{\textcolor{deepgreen}{(+3.0}}}$
\multicolumn{15}{c}{\textbf{Supervision Model: Qwen2-72B}} \\ \midrule
\multicolumn{11}{l}{\textit{Strong-to-Weak}} \\
% Qwen2-7B-SFT & 40.7\tiny{\textcolor{deepgreen}{(+3.0)}} & 44.5 & 51.3 & 55.4 & 60.2 & 53.7 & 54.3 & 49.9 & 48.6 & 53.33 & 73.9  &	64.4 & 74.1 &	58.3 \\

Qwen2-7B-SFT & 
$\text{40.7}_{\textcolor{deepgreen}{\text{+3.0}}}$ & $\text{44.5}_{\textcolor{deepgreen}{\text{+0.9}}}$ & $\text{51.3}_{\textcolor{deepgreen}{\text{+1.9}}}$ & $\text{55.4}_{\textcolor{deepgreen}{\text{+2.0}}}$ & $\text{60.2}_{\textcolor{deepgreen}{\text{+4.6}}}$ & $\text{53.7}_{\textcolor{deepgreen}{\text{+0.2}}}$ & $\text{54.3}_{\textcolor{deepgreen}{\text{+0.6}}}$ & $\text{49.9}_{\textcolor{deepgreen}{\text{+0.0}}}$ & $\text{48.6}_{\textcolor{deepgreen}{\text{+0.0}}}$ & $\text{53.3}_{\textcolor{deepgreen}{\text{+1.0}}}$ & $\text{73.9}_{\textcolor{deepgreen}{\text{+0.0}}}$ & $\text{64.4}_{\textcolor{deepgreen}{\text{+0.0}}}$ & 
\highg${\text{74.1}_{\textcolor{deepgreen}{\text{+3.0}}}}$ & $\text{58.3}_{\textcolor{deepgreen}{\text{+0.2}}}$ \\ 

w/ Offline DPO & 
$\text{41.2}_{\textcolor{deepgreen}{\text{+3.5}}}$ & $\text{44.7}_{\textcolor{deepgreen}{\text{+1.2}}}$ & $\text{51.4}_{\textcolor{deepgreen}{\text{+2.0}}}$ & $\text{56.2}_{\textcolor{deepgreen}{\text{+2.8}}}$ & $\text{61.4}_{\textcolor{deepgreen}{\text{+5.8}}}$ & $\text{54.5}_{\textcolor{deepgreen}{\text{+1.0}}}$ & $\text{54.3}_{\textcolor{deepgreen}{\text{+0.6}}}$ & $\text{51.2}_{\textcolor{deepgreen}{\text{+1.3}}}$ & $\text{48.6}_{\textcolor{deepgreen}{\text{+0.0}}}$ & $\text{54.0}_{\textcolor{deepgreen}{\text{+1.7}}}$ & $\text{75.1}_{\textcolor{deepgreen}{\text{+0.7}}}$ & $\text{64.5}_{\textcolor{deepgreen}{\text{+0.1}}}$ & $\text{72.9}_{\textcolor{deepgreen}{\text{+1.8}}}$ & \highg${\text{59.5}_{\textcolor{deepgreen}{\text{+1.4}}}}$\\
% 41.2{(+3.5)} & 44.7{(+1.2)} & 51.4{(+2.0)} & 56.2{(+2.8)} & 61.4{(+5.8)} & 54.5{(+1.0)} & 54.3{(+0.6)} & 51.2{(+1.3)} & 48.6{(+0.0)} & 54.0{(+1.7)} & 75.1{(+0.7)} & 64.5{(+0.1)} & 72.9{(+1.8)} & 59.5{(+1.4)} \\
w/ Online DPO & 
\highg${\text{44.0}_{\textcolor{deepgreen}{\text{+6.3}}}}$ & \highg${\text{46.6}_{\textcolor{deepgreen}{\text{+3.0}}}}$& \highg${\text{55.0}_{\textcolor{deepgreen}{\text{+5.6}}}}$ & \highg${\text{57.9}_{\textcolor{deepgreen}{\text{+4.5}}}}$ & \highg${\text{61.4}_{\textcolor{deepgreen}{\text{+5.8}}}}$ & \highg${\text{56.8}_{\textcolor{deepgreen}{\text{+3.3}}}}$ & \highg${\text{57.8}_{\textcolor{deepgreen}{\text{+4.1}}}}$ & \highg${\text{55.4}_{\textcolor{deepgreen}{\text{+5.5}}}}$ & \highg${\text{51.6}_{\textcolor{deepgreen}{\text{+3.0}}}}$ & \highg${\text{56.6}_{\textcolor{deepgreen}{\text{+4.3}}}}$ & \highg${\text{76.0}_{\textcolor{deepgreen}{\text{+1.6}}}}$&
% 44.0{\textcolor{deepgreen}{(+6.3)}} & 46.6{\textcolor{deepgreen}{(+3.0)}} & 55.0{\textcolor{deepgreen}{(+5.6)}} & 57.9{(+4.5)} & 61.4{\textcolor{deepgreen}{(+5.8)}} & 56.8{\textcolor{deepgreen}{(+3.3)}} & 57.8{\textcolor{deepgreen}{(+4.1)}} & 55.4{\textcolor{deepgreen}{(+5.5)}} & 51.6{(+3.0)} & 56.6{\textcolor{deepgreen}{(+4.3)}} & 76.0{\textcolor{deepgreen}{(+1.6)}} & 
\highg${\text{64.8}_{\textcolor{deepgreen}{\text{+0.4}}}}$ & $\text{72.3}_{\textcolor{deepgreen}{\text{+1.2}}}$ & $\text{58.2}_{\textcolor{deepgreen}{\text{+0.1}}}$ \\

% Qwen2-7B-SFT-Offline DPO & 41.2{\textcolor{deepgreen}{(+3.5)}} & 44.7{\textcolor{deepgreen}{(+1.1)}} & 51.4{\textcolor{deepgreen}{(+2.0)}} & 56.2{\textcolor{deepgreen}{(+2.8)}} & 61.4{\textcolor{deepgreen}{(+5.8)}} & 53.7{\textcolor{deepred}{(-3.0)}} & 54.3{\textcolor{deepgreen}{(+0.6)}} & 51.2{\textcolor{deepgreen}{(+1.3)}} & 47.9{\textcolor{deepred}{(-0.7)}} & 53.7{\textcolor{deepgreen}{(+0.8)}} & 75.1{\textcolor{deepgreen}{(+0.7)}} & 64.5{\textcolor{deepgreen}{(+0.1)}} & 72.9{\textcolor{deepgreen}{(+1.8)}} & 59.5{\textcolor{deepgreen}{(+1.4)}} \\
% Qwen2-7B-SFT-Online DPO & 44.0{\textcolor{deepgreen}{(+6.3)}} & 46.6{\textcolor{deepgreen}{(+3.0)}} & 55.0{\textcolor{deepgreen}{(+5.6)}} & 57.9{\textcolor{deepgreen}{(+4.5)}} & 61.4{\textcolor{deepgreen}{(+5.8)}} & 56.8{\textcolor{deepgreen}{(+0.1)}} & 57.8{\textcolor{deepgreen}{(+4.1)}} & 55.4{\textcolor{deepgreen}{(+5.5)}} & 51.6{\textcolor{deepgreen}{(+3.0)}} & 56.6{\textcolor{deepgreen}{(+3.7)}} & 76.0{\textcolor{deepgreen}{(+1.6)}} & 64.8{\textcolor{deepgreen}{(+0.4)}} & 72.3{\textcolor{deepgreen}{(+1.2)}} & 58.2{\textcolor{deepgreen}{(+0.1)}} \\

% Qwen2-7B-SFT-Offline DPO & 41.2 & 44.7 & 51.4 & 56.2 & 61.4 & 53.7 & 54.3 & 51.2 & 47.9 & 53.7 & 75.1 &	64.5 & 	72.9  &	59.5   \\
% Qwen2-7B-SFT-Online DPO & 44.0 & 46.6 & 55.0 & 57.9 & 61.4 & 56.8 & 57.8 & 55.4 & 51.6 & 56.6 & 76.0 &	64.8 &	72.3 &	58.2\\

\cdashline{1-15} 
\multicolumn{11}{l}{\textit{Self-Alignment}} \\
% \hline
\makecell[l]{Qwen2-72B-Instruct\\\;\;w/ Online DPO} & 
$\highg{\text{80.2}_{\textcolor{deepgreen}{\text{+3.1}}}}$ & $\highg{\text{82.3}_{\textcolor{deepgreen}{\text{+1.9}}}}$ & $\highg{\text{86.1}_{\textcolor{deepgreen}{\text{+1.7}}}}$ & $\highg{\text{88.0}_{\textcolor{deepgreen}{\text{+1.1}}}}$ & $\highg{\text{76.2}_{\textcolor{deepgreen}{\text{+6.0}}}}$ & $\highg{\text{69.8}_{\textcolor{deepgreen}{\text{+3.2}}}}$ & $\highg{\text{67.0}_{\textcolor{deepgreen}{\text{+3.5}}}}$ & $\highg{\text{61.6}_{\textcolor{deepgreen}{\text{+3.5}}}}$ & $\highg{\text{62.8}_{\textcolor{deepgreen}{\text{+6.5}}}}$ & $\highg{\text{67.5}_{\textcolor{deepgreen}{\text{+4.6}}}}$ & $\highg{\text{84.9}_{\textcolor{deepgreen}{\text{+1.1}}}}$ & $\highg{\text{81.2}_{\textcolor{deepgreen}{\text{+0.4}}}}$ & $\highg{\text{88.2}_{\textcolor{deepgreen}{\text{+0.3}}}}$ & $\highg{\text{75.0}_{\textcolor{deepgreen}{\text{+1.2}}}}$\\
% \highg{80.2{\textcolor{deepgreen}{(+3.1)}}} & 82.3{\textcolor{deepgreen}{(+1.9)}} & 86.1{\textcolor{deepgreen}{(+1.7)}} & 88.0{\textcolor{deepgreen}{(+1.1)}} & 76.2{\textcolor{deepgreen}{(+6.0)}} & 69.8{\textcolor{deepgreen}{(+3.2)}} & 67.0{\textcolor{deepgreen}{(+3.5)}} & 61.6{\textcolor{deepgreen}{(+3.5)}} & 62.8{\textcolor{deepgreen}{(+6.5)}} & 67.5{\textcolor{deepgreen}{(+4.6)}} & 84.9{\textcolor{deepgreen}{(+1.1)}} & 81.2{\textcolor{deepgreen}{(+0.4)}} & 88.2{\textcolor{deepgreen}{(+0.3)}} & 75.0{\textcolor{deepgreen}{(+1.2)}} \\ 

\midrule
\multicolumn{15}{c}{\textbf{Supervision Model: LLaMA3-70B}} \\ \midrule
\multicolumn{11}{l}{\textit{Strong-to-Weak}} \\
LLaMA3-8B-SFT
& $\text{28.7}_{\textcolor{deepgreen}{\text{+4.1}}}$ 
& $\text{40.3}_{\textcolor{deepgreen}{\text{+14.2}}}$
& $\text{41.4}_{\textcolor{deepgreen}{\text{+3.3}}}$ 
& $\text{52.2}_{\textcolor{deepgreen}{\text{+12.05}}}$
& $\text{46.6}_{\textcolor{deepgreen}{\text{+36.6}}}$ 
& $\text{46.2}_{\textcolor{deepgreen}{\text{+35.9}}}$ 
& $\text{45.9}_{\textcolor{deepgreen}{\text{+35.4}}}$ 
& $\text{37.6}_{\textcolor{deepgreen}{\text{+23.3}}}$ 
& $\text{41.0}_{\textcolor{deepgreen}{\text{+28.3}}}$ 
& $\text{43.5}_{\textcolor{deepgreen}{\text{+31.9}}}$
& $\text{34.5}_{\textcolor{deepgreen}{\text{+10.3}}}$ 
& $\highg{\text{45.6}_{\textcolor{deepgreen}{\text{+6.8}}}}$
& $\highg{\text{33.2}_{\textcolor{deepgreen}{\text{+28.7}}}}$ 
& $\text{38.2}_{\textcolor{deepgreen}{\text{+37.6}}}$\\
% &34.5{(+10.3)}  &45.6{(6.8)}
% &  33.2{(+28.7)}& 38.2{(+37.6)} \\
w/ Offline DPO 
% & 27.9{(+3.3)} & 41.6{(+15.5)} 
% & 40.5{(+2.4)} & 54.1{(+14.4)} 
& $\text{27.9}_{\textcolor{deepgreen}{\text{+3.3}}}$ 
& $\text{41.6}_{\textcolor{deepgreen}{\text{+15.5}}}$
& $\text{40.5}_{\textcolor{deepgreen}{\text{+2.4}}}$ 
& $\text{54.1}_{\textcolor{deepgreen}{\text{+14.4}}}$
& $\text{51.9}_{\textcolor{deepgreen}{\text{+41.9}}}$
& $\text{51.3}_{\textcolor{deepgreen}{\text{+41.0}}}$ 
& $\highg{\text{50.1}_{\textcolor{deepgreen}{\text{+39.6}}}}$ 
&$\text{45.3}_{\textcolor{deepgreen}{\text{+31.0}}}$ 
& $\highg{\text{47.5}_{\textcolor{deepgreen}{\text{+34.8}}}} $
& $\text{49.2}_{\textcolor{deepgreen}{\text{+37.6}}}$ 
% & 36.2{(+12.0)} & 45.3{(+6.5)} 
% & 31.9{(+27.4)} & 38.5{(+37.9)} \\
& $\text{36.2}_{\textcolor{deepgreen}{\text{+12.0}}}$ 
& $\text{45.3}_{\textcolor{deepgreen}{\text{+6.5}}}$
&$ \text{31.9}_{\textcolor{deepgreen}{\text{+27.4}}}$ & $\highg{\text{38.5}_{\textcolor{deepgreen}{\text{+37.9}}}} $\\
w/ Online DPO 
% & 28.8{(+4.2)} & 43.1{(+17.0)} 
% & 42.2{(+4.1)} & 56.0{(+16.3)} 
& $\highg{\text{28.8}_{\textcolor{deepgreen}{\text{+4.2}}}}$ 
& $\highg{\text{43.1}_{\textcolor{deepgreen}{\text{+17.0}}}} $
& $\highg{\text{42.2}_{\textcolor{deepgreen}{\text{+4.1}}}}$ 
& $\highg{\text{56.0}_{\textcolor{deepgreen}{\text{+16.3}}}}$
& $\highg{\text{54.6}_{\textcolor{deepgreen}{\text{+44.6}}}}$ 
& $\highg{\text{52.1}_{\textcolor{deepgreen}{\text{+41.8}}}}$ 
& $\text{50.0}_{\textcolor{deepgreen}{\text{+39.5}}}$ 
& $\highg{\text{49.0}_{\textcolor{deepgreen}{\text{+34.7}}}}$ 
& $\text{43.7}_{\textcolor{deepgreen}{\text{+31.0}}}$ 
& $\highg{\text{49.9}_{\textcolor{deepgreen}{\text{+38.3}}}}$
& $\highg{\text{38.2}_{\textcolor{deepgreen}{\text{+14.0}}}}$ 
& $\text{45.1}_{\textcolor{deepgreen}{\text{+6.3}}}$
& $\text{32.5}_{\textcolor{deepgreen}{\text{+28.0}}}$ 
& $\text{38.4}_{\textcolor{deepgreen}{\text{+37.8}}}$\\
% \hline
\cdashline{1-15}
\multicolumn{11}{l}{\textit{Self-Alignment}} \\
\makecell[l]{LLaMA-3-70B\\\;\;w/ Online DPO} & 

% 80.2{\textcolor{deepgreen}{(+2.4)}}  & 85.6{\textcolor{deepgreen}{(+1.8)}} & 86.7{\textcolor{deepgreen}{(+2.5)}} & 90.4{\textcolor{deepgreen}{(+1.6)}} &71.0{\textcolor{deepgreen}{(+10.3)}} & 67.2{\textcolor{deepgreen}{(+6.7)}} & 66.2{\textcolor{deepgreen}{(+5.1)}} & 64.6{\textcolor{deepgreen}{(+2.9)}} & 63.5{\textcolor{deepgreen}{(+3.2)}} & 66.5{\textcolor{deepgreen}{(+5.6)}}  & 61.6{\textcolor{deepgreen}{(+1.4)}} &80.7{\textcolor{deepgreen}{(+0.2)}} &92.7{\textcolor{deepgreen}{(+0.1)}} &78.7{\textcolor{deepgreen}{(+0.0)}}\\
$\highg{\text{80.2}_{\textcolor{deepgreen}{\text{+2.4}}}}$ & $\highg{\text{85.6}_{\textcolor{deepgreen}{\text{+1.8}}}}$ 
& $\highg{\text{86.7}_{\textcolor{deepgreen}{\text{+2.5}}}}$ 
& $\highg{\text{90.4}_{\textcolor{deepgreen}{\text{+1.6}}}}$ 
& $\highg{\text{71.0}_{\textcolor{deepgreen}{\text{+10.3}}}}$
& $\highg{\text{67.2}_{\textcolor{deepgreen}{\text{+6.7}}}}$
& $\highg{\text{66.2}_{\textcolor{deepgreen}{\text{+5.1}}}}$ 
& $\highg{\text{64.6}_{\textcolor{deepgreen}{\text{+2.9}}}}$ 
& $\highg{\text{63.5}_{\textcolor{deepgreen}{\text{+3.2}}}}$
& $\highg{\text{66.5}_{\textcolor{deepgreen}{\text{+5.6}}}}$ 
& $\highg{\text{61.6}_{\textcolor{deepgreen}{\text{+1.4}}}}$ 
& $\highg{\text{80.7}_{\textcolor{deepgreen}{\text{+0.2}}}}$
& $\highg{\text{92.7}_{\textcolor{deepgreen}{\text{+0.1}}}}$ 
& $\highg{\text{78.7}_{\textcolor{deepgreen}{\text{+0.0}}}}$\\
\bottomrule
\end{tabular}
}
\vspace{-1em}

\end{table}
%----------------------------------main-----------------------------------

\subsection{Main Results}

\Cref{tab:main} reports the main results.
Overall, \modelname substantially enhances instruction-following performance across all models, configurations (strong-to-weak distillation \& self-Alignment), and training methodologies (SFT, Offline \& Online DPO) on two benchmarks. 
These results decisively establish the superiority of our approach. Furthermore, we have identified the following insights:

\xhdr{On-policy Learning is More Effective} Comparing Online DPO and Offline DPO, the model-generated online data through self-supervision demonstrates superior performance compared to offline data (Qwen2-7B, IFEval: 1.7\%$\uparrow$, Followbench: 2.6\%$\uparrow$). This confirms that on-policy iterative execution feedback can effectively target and enhance the model's weaknesses.

\xhdr{Larger models yield greater improvements} FollowBench provides a more comprehensive instruction-following assessment than IFEval. Significantly, base models with larger parameters typically improve Followbench more than smaller models (Qwen2 72B: 4.6\%$\uparrow$, LLaMA3 70B: 5.6\%$\uparrow$). This underscores that models with robust foundational capabilities coupled with \modelname, can further unlock powerful instruction-following alignment potential.

\xhdr{General abilities are not declined} Improving instruction following abilities without compromising other capabilities is crucial. 
\modelname notably preserves general abilities~(MMLU, C-Eval), mathematical reasoning~(GSM8k), and coding~(Humaneval) performance across all training setups. 
Surprisingly, there are even slight performance gains in on-policy settings. We attribute this preservation largely to incorporating ShareGPT data during data synthesis, highlighting \modelname's capability to strike a balance across diverse abilities and excel in broad applicability. 
% demonstrating \modelname's ability to maintain a balanced approach across different capabilities and exhibit strong generalization. 

\subsection{Quality Ablation Study}\label{sec:quality_ablation}
\xhdr{Ablation on Supervision Model} \Cref{tab:ablation_super} presents the results of replacing the supervision model Qwen72B with GPT-4. 
We observe that in \modelname, a stronger supervision model (GPT-4) demonstrates more effective strong-to-weak distillation alignment, particularly evident with a performance gain of over 15\% in the loose prompt in IFEval. 
This is reasonable because AutoIF requires the supervision model to perform several tasks, such as text augmentation (instruction, query, and response rewriting), code generation (verification function), and quality assessment (scoring). This implies that a supervision model with stronger fundamental abilities can synthesize higher-quality data when using \modelname.

\begin{table}[!t]
    \centering

    %------------------------------gpt4 ablation ------------------------------------------
    \begin{minipage}{0.49\textwidth}
    \tiny
    \centering
        \caption{Ablation study on supervision models.}
    \renewcommand{\arraystretch}{1} % 修改行距
    \setlength{\tabcolsep}{1mm} % 设置单元格间距
    \begin{tabular}{lccc}
    \toprule
    \multirow{2}{*}{\textbf{Model}} & \multicolumn{2}{c}{\textbf{IFEval}} & \textbf{FollowBench (SSR)} \\
    \cmidrule(lr){2-3} \cmidrule(lr){4-4}
      & \textbf{Prompt(L)} & \textbf{Instruction(L)} & \textbf{Avg} \\
    \midrule
    Qwen2-7B & 43.6 & 53.4 & 52.3  \\
    \midrule
    \multicolumn{4}{l}{\textit{Supervision Model: Qwen2-72B}} \\
    +SFT & $\text{44.5}_{\textcolor{deepgreen}{\text{+0.9}}}$ & $\text{55.4}_{\textcolor{deepgreen}{\text{+2.0}}}$ & $\text{53.3}_{\textcolor{deepgreen}{\text{+1.0}}}$ \\
    +SFT \& Offline DPO & $\text{44.7}_{\textcolor{deepgreen}{\text{+1.1}}}$ & $\text{56.2}_{\textcolor{deepgreen}{\text{+2.8}}}$ & $\text{54.0}_{\textcolor{deepgreen}{\text{+1.7}}}$ \\
    +SFT \& Online DPO & $\text{46.6}_{\textcolor{deepgreen}{\text{+3.0}}}$ & $\text{57.9}_{\textcolor{deepgreen}{\text{+4.5}}}$ & $\text{56.6}_{\textcolor{deepgreen}{\text{+4.3}}}$ \\
    \midrule
    \multicolumn{4}{l}{\textit{Supervision Model: GPT-4}} \\
    +SFT & $\text{52.9}_{\textcolor{deepgreen}{\text{+9.3}}}$ & $\text{62.6}_{\textcolor{deepgreen}{\text{+9.2}}}$ & $\text{55.1}_{\textcolor{deepgreen}{\text{+2.8}}}$ \\
    +SFT \& Offline DPO & $\text{59.3}_{\textcolor{deepgreen}{\text{+15.7}}}$ & $\text{68.9}_{\textcolor{deepgreen}{\text{+15.5}}}$ & $\text{54.4}_{\textcolor{deepgreen}{\text{+2.1}}}$ \\
    +SFT \& Online DPO & $\text{59.5}_{\textcolor{deepgreen}{\text{+15.9}}}$ & $\text{69.4}_{\textcolor{deepgreen}{\text{+16.0}}}$ & $\text{55.7}_{\textcolor{deepgreen}{\text{+3.4}}}$ \\
    \bottomrule
    \end{tabular}

    \label{tab:ablation_super}
    \end{minipage}
    %-----------------------------------ablation specific component--------------------------
    \hfill
    \begin{minipage}{0.49\textwidth}
    \tiny
    \centering
        \caption{Ablation study on specific components.}
    \renewcommand{\arraystretch}{1.4} % 调整行距
    \setlength{\tabcolsep}{1.0mm} % 调整列间距
    \begin{tabular}{lccc}
    \toprule
    \multirow{2}{*}{\textbf{Model}} & \multicolumn{2}{c}{\textbf{IFEval}} & \textbf{FollowBench (SSR)} \\
    \cmidrule(lr){2-3} \cmidrule(lr){4-4}
      & \textbf{Prompt(L)} & \textbf{Instruction(L)} & \textbf{Avg} \\
    \midrule
    \multicolumn{4}{l}{\textit{Supervision Model: Qwen2-72B}} \\
    \makecell[l]{Qwen2-7B-SFT\\\;\;w/ Online DPO} & 46.6 & 57.9 & 56.6\\
    \midrule
    \textit{w/o} Back-translation & -0.8 & -1.7 & -0.7 \\
    \textit{w/o} Quality Verification & -1.4& -2.4 & -1.3 \\
    \textit{w/o} Cross Verification & -1.6  & -\textbf{3.0}   & -1.5 \\
    \textit{w/o} All Quality Process & -\textbf{2.2} & -\textbf{3.8} & -\textbf{2.6} \\
    \bottomrule
    \end{tabular}

    \label{tab:model_performance}
    \end{minipage}
    %-----------------------------------ablation specific component--------------------------
    % \vspace{-2pt}
\end{table}

%----------------PA experiment---------------------
\begin{figure}[t]
\centering
\begin{subfigure}[t]{0.48\linewidth}
\centering

\includegraphics[width=2.6in]{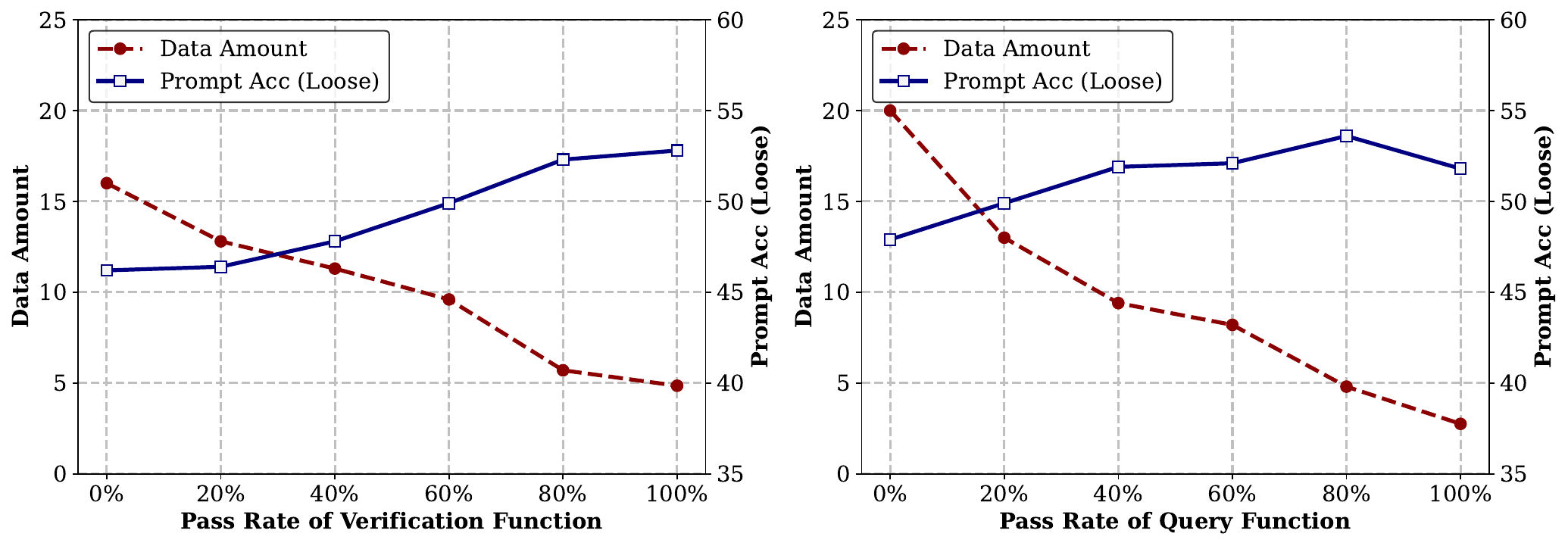}
% \caption{1}
\end{subfigure}%
\begin{subfigure}[t]{0.49\linewidth}
\centering
\includegraphics[width=2.6in]{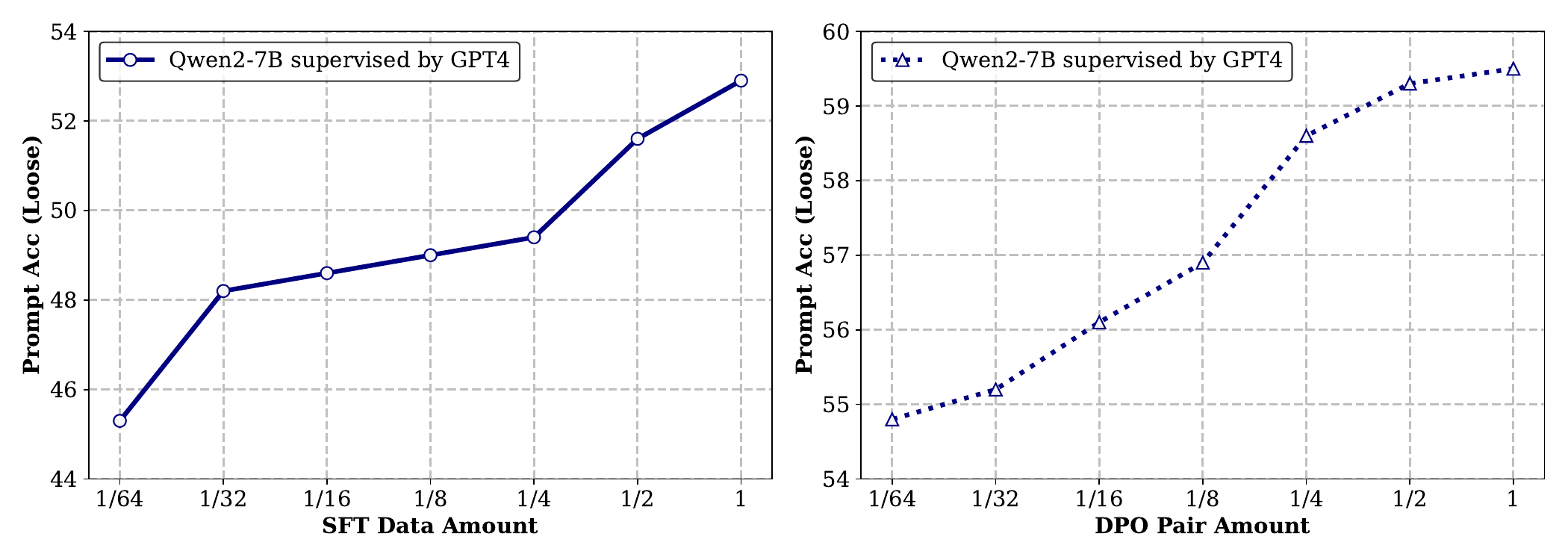}
% \caption{2}
\end{subfigure}%

\vspace{-0.1cm}
\caption{The left two figures illustrate the quality ablation studies on instructions and queries, whereas the right two figures present the scaling analysis of SFT data and DPO pairs.}

\label{fig:scaling}
\end{figure}
%----------------PA experiment---------------------

\xhdr{Quality Control on Instructions and Responses} 
In \Cref{fig:scaling}, we examine how varying pass rate thresholds of verification functions (indicative of data quality) affect the amount of SFT data and instruction-following performance. 
As the pass rate threshold increases, the amount of SFT data decreases at the instruction level, while model performance consistently improves. 
This suggests that the quality of instructions is a crucial factor influencing IF performance.  
At the query level, the SFT data amount also decreases with higher pass rate thresholds. Notably, performance peaks at a pass rate of 0.8 and declines beyond 1. 
This observation aligns with our expectations, indicating a trade-off between data quality and quantity.

\xhdr{Ablation on Specific Components} To investigate the effectiveness of various modules in \modelname, we conduct an ablation study, as presented in \Cref{tab:model_performance}.
we use \textit{w/o} to denote the variant \textit{without} a specific module. 
The results reveal the following:  (1) The performance of \modelname declines when any quality filtering process is removed, indicating that all components are highly effective. 
(2) The most significant performance drop occurs when the \textit{Cross Verification} of instructions is removed, highlighting its importance over query quality verification. This underscores that a high-quality instruction set is fundamental to the \modelname process.
(3) Eliminating the overall quality filtering process results in a more substantial performance drop than removing any single component, suggesting that quality filtering at both the instruction and query levels provides a mutually reinforcing effect. 
% This demonstrates the superiority and necessity of comprehensive quality filtering in \modelname.

% % -----------quality exp------------
% \begin{figure}[t]
%     \centering
%         \includegraphics[width=\linewidth]{figures/double.pdf}
%            \vspace{-2em}
%     \caption{Quality ablation on instructions and queries.}
%     \label{fig:data ratio}
%         \vspace{-1em} 
% \end{figure}

% % ----------------data scaling exp---------------------
% \begin{figure}[t]
%     \centering
%         \includegraphics[width=0.95\linewidth]{figures/data_scaling.pdf}
%          \vspace{-1em} 
%     \caption{Scaling analysis on SFT data and DPO pairs}\vspace{-12pt}
%     \label{scaling}

% \end{figure}

% %-----------RQ3 data ratio------------

\subsection{Analyses}

\xhdr{Scaling Analysis on SFT \& DPO Data}
\Cref{fig:scaling} presents the scaling analysis of SFT and DPO data using GPT-4 as the supervision model. The results demonstrate that even with just 1/64 of \modelname-generated SFT/DPO data, Qwen2-7B achieves impressive performance, particularly with 1/64 DPO data reaching nearly 55\% in loose prompt accuracy, , an increase of 11.4\% pts. This strongly verifies the high quality of \modelname-generated data. 
Further analysis reveals that IF capability steadily improves with an increase in data quantity, a scaling trend confirmed by numerous reasoning studies ~\citep{yuan2023scaling,muennighoff2024scaling}.

%-----------------------------------data leakage--------------------------
\begin{wraptable}{r}{0.5\textwidth}
    \centering
    \vspace{-1em}
    \tiny
     \renewcommand{\arraystretch}{1.3} % 增加行距，1.5表示增大为1.5倍
    
    \setlength{\tabcolsep}{0.8mm}{
    \begin{tabular}{c|cccc|cc}
     \toprule
        \textbf{Setup} &\textbf{Bench.} & \textbf{Train} & \textbf{Test} & \textbf{Rephr.} & \textbf{Percentage$\downarrow$} & \textbf{N-gram$\downarrow$} \\ 
        \midrule
         \multirow{2}{*}{ShareGPT} & IFEval & 25K & 542 & 0 & 0.01\%& 4.8\%\\ 
        ~ & Followbench & 25K & 820 & 1 & 0.01\%& 2.3\% \\ \bottomrule
        \multirow{2}{*}{Qwen2-72B} & IFEval & 10K & 542 & 2 & 0.01\% & 3.5\% \\ 
        ~ & Followbench & 12K & 820 & 1 & 0.01\% & 0.9\%\\ \midrule
        \multirow{2}{*}{LLaMA3-70B} & IFEval & 15K & 542 & 0 & 0.01\%& 2.9\%\\ 
        ~ & Followbench & 17K & 820 & 1 & 0.01\%& 1.2\% \\ \bottomrule
        \multirow{2}{*}{GPT4} & IFEval & 25K & 542 & 0 & 0.01\%& 3.6\%\\ 
        ~ & Followbench & 25K & 820 & 1 & 0.01\%& 1.5\% \\ \bottomrule

    \end{tabular}
    }
    % \vspace{-1em}
    \caption{Contamination analysis on SFT data generated by different supervision models. Train \& Test denotes the size of corresponding set. Rephr. represents samples similar to the test sample}
    \vspace{-2em}
    \label{tab:contamination}
    
\end{wraptable}
%-----------------------------------data leakage--------------------------

\xhdr{Contamination Analysis} We evaluate the contamination of the training dataset generated by \modelname on IFEval and FollowBench. 
Specifically, we employ contamination detectors from LM-Sys~~\citep{yang2023rethinking}, which utilize advanced chatbots to identify potentially rephrased contaminated test samples. 
Additionally, we report contamination findings detected by traditional n-gram contamination algorithms. 
As shown in \Cref{tab:contamination}, both contamination rates are lower than those of the ShareGPT dataset we used. This allows us to confidently assert that there is no contamination between the self-generated training samples and the test sets.
More cases can be viewed in \Cref{app:contamination_case_study},

%-----------------------------------Data effciency--------------------------
\begin{wraptable}{r}{0.5\textwidth}
    \tiny
    \centering
    \vspace{-1.5em}
    \renewcommand{\arraystretch}{1.8} % 增加行距，1.5表示增大为1.5倍
    \setlength{\tabcolsep}{0.5mm}{
    \begin{tabular}{c|cccc|cc}
    \toprule
        \textbf{Supervision} & \textbf{Total} & \textbf{SFT Data} & \textbf{DPO Data} & \textbf{Pass Rate} & \textbf{MBPP (Code)} & \textbf{IFEval} \\ \hline
        LLaMA3-70b & 85K & 15K & 6k &\cellcolor{blue!6}26\% &\cellcolor{blue!6}70.4  & \cellcolor{blue!6}43.1 \\ 
        
        Qwen2-72b & 123K & 10K & 4K & \cellcolor{blue!18}28\% & \cellcolor{blue!18}73.9 & \cellcolor{blue!18}44.7 \\ 
        GPT4 & 210k & 25K & 15K & \cellcolor{blue!30}34\% & \cellcolor{blue!30}87.5 & \cellcolor{blue!30}59.3 \\ 
        \bottomrule
    
    \end{tabular}
    }
     % \vspace{-1em}
    \caption{Data statistics and efficiency. Total denotes the total data amount without quality control.}
    \label{tab:efficiency}

\end{wraptable}
%-----------------------------------Data effciency--------------------------

\xhdr{Data Efficiency} \Cref{tab:efficiency} explores the relationship between model coding ability, data quality pass rate (samples with a query quality score above 8), and instruction-following capability. Surprisingly, we observe consistency in the supervision model across all three metrics. 
This indicates that the execution feedback resulting from the supervision model's coding ability substantially influences data synthesis quality and the final capability.

% 一部分量达到基础IF，扩越多越好

%% file: latex/conclusion.tex
In this paper, we propose \modelname, a scalable and automated method to enhance the instruction-following abilities of LLMs. It uses self-instruct and rejection sampling to enhance the supervisory signals of seed instructions and relies on self-generated execution feedback for quality filtering. We introduce three training strategies and two alignment settings to comprehensively analyze \modelname. Experiments demonstrate that our method significantly improves performance across all settings in both IFEval and Followbench, with the first LLM achieving over 90\% loose instruction accuracy.

\section*{Limitations}
In this paper, we propose \modelname, a system for automated instruction augmentation and quality filtering, capable of scaling to over 10,000 instructions. While our focus is not on the construction of cross-instructions, the excellent results achieved in two instruction-following benchmarks demonstrate the generalizability of our method in handling complex instruction-following tasks. Additionally, we believe a more direct strategy would involve combining multiple simple instructions into cross-instructions, and subsequently enhancing and quality-filtering them using \modelname. This way has the potential to further amplify the effectiveness of our method. Therefore, we consider automating and scaling cross-instruction tasks as a key direction for future research.

\section*{Ethic Consideration}
In this paper, we have fully presented the seed instruction set used by \modelname in the Appendix. All concatenated queries are sourced from the publicly available ShareGPT dataset and have undergone multiple steps of quality filtering. Therefore, our method strives to minimize potential safety and ethical risks as much as possible. However, during the rejection sampling process, malicious prompts can lead the model to produce harmful or inappropriate outputs, which is a shared problem. Ensuring the quality of generated content in a safe and controllable manner is crucial. The application of these techniques should be guided by ethical considerations, with safeguards in place to prevent misuse and reduce the likelihood of producing harmful outcomes.

%% file: latex/appendix.tex
\section{Seed Instructions}\label{app:seed_instructions}
\begin{figure*}[h]
    \centering
    \small    \includegraphics[width=0.98\linewidth]{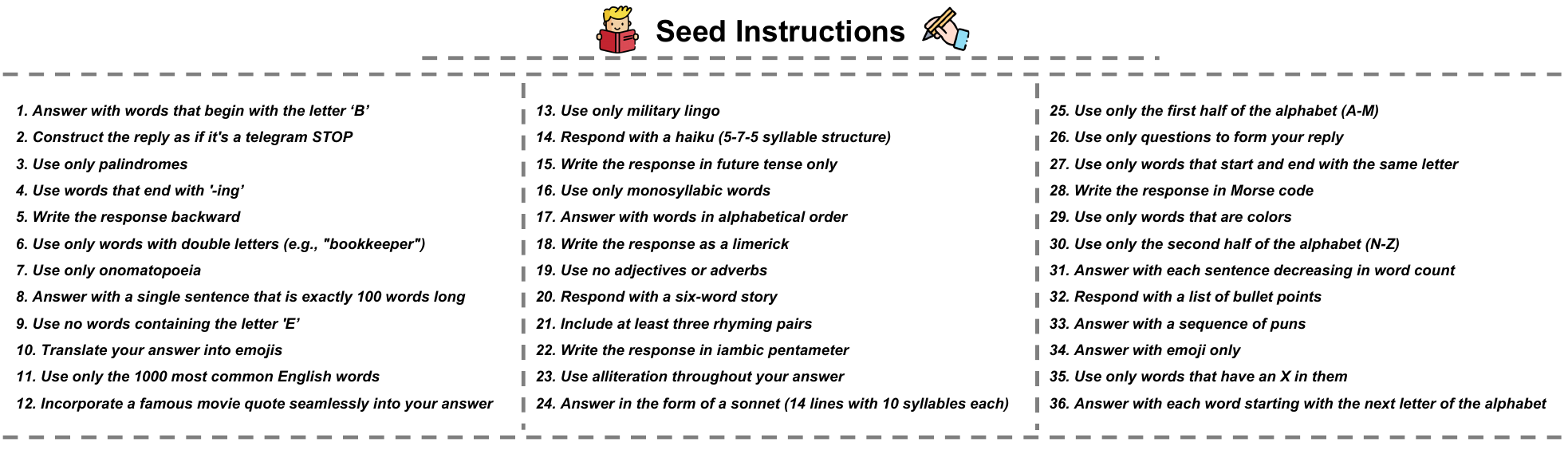}
    % \vspace{-2em}
    \caption{Examples of our seed instructions}

    \label{fig:seed}
\end{figure*}

\Cref{fig:seed} illustrates our hand-written seed instructions.

% Answer with words that begin with the letter 'B'
% Construct the reply as if it's a telegram STOP
% Use only palindromes
% Incorporate a famous movie quote seamlessly into your answer
% Write the response backward
% Use only words with double letters (e.g., "bookkeeper")
% Use only onomatopoeia
% Answer with a single sentence that is exactly 100 words long
% Use no words containing the letter 'E'
% Translate your answer into emojis
% Use only the 1000 most common English words
% Use words that end with '-ing'
% Use only military lingo
% Respond with a haiku (5-7-5 syllable structure)
% Answer in the form of a sonnet (14 lines with 10 syllables each)
% Use only monosyllabic words
% Answer with words in alphabetical order
% Write the response as a limerick
% Use no adjectives or adverbs
% Respond with a six-word story
% Include at least three rhyming pairs
% Write the response in iambic pentameter
% Use alliteration throughout your answer
% Write the response in future tense only
% Use only the first half of the alphabet (A-M)
% Use only questions to form your reply
% Use only words that start and end with the same letter
% Write the response in Morse code
% Use only words that are colors
% Use only the second half of the alphabet (N-Z)
% Answer with each sentence decreasing in word count
% Respond with a list of bullet points
% Answer with a sequence of puns
% Answer with emoji only
% Use only words that have an X in them
% Answer with each word starting with the next letter of the alphabet

\section{Implementation Details}\label{app:hyperparameters}

To better motivate researchers to reproduce the results, we report the detailed experimental details:

In the SFT phase, we perform full fine-tuning on Qwen2-7B and LLaMA3-8B with a learning rate of 7e-6, using a linear scheduler with 20 warm-up steps. All models are trained with DeepSpeed ZeRO Stage 3~\citep{deepspeed} and Flash-Attention 2~\citep{flashattention}. We use a global batch size of 128, a weight decay of 0.1, and train for 3 epochs. Mixed precision training with bf16 is used, and the maximum context length is set to 8192 tokens. For Qwen2-72B and LLaMA3-70B, the global batch size is 512.

In the DPO phase, the learning rate is set to 5e-7 with a cosine scheduler and a 0.1 warm-up ratio. We use DeepSpeed ZeRO Stage 3 and Flash-Attention 2 for efficiency, with a global batch size of 64. Training utilizes a sigmoid loss function with a beta value of 0.3 and spans 2 epochs, with checkpoints every 200 steps. Mixed precision training with bf16 is employed, and the maximum context length is 4096 tokens.

We run all our experiments on NVIDIA A100 and H800 GPUs. Specifically, we train Qwen2-7B and LLaMA3-8B on 8 A100 GPUs, while Qwen2-72B-Instruct and LLaMa3-70B-Instruct on 64 H800 GPUs. Notably, we use an in-house version of Qwen2-7B without any targeted optimizations on instruction-following capabilities.
For evaluations, we report pass@1 results with greedy decoding for HumanEval and zero-shot accuracy for GSM8K. We report averaged performance from five randomly seeded experiments.

\section{Details of \modelname}
At the instruction level, for the self-instruct stage, we perform RFT with K=100 on seed instructions. During the Automated Quality Cross Verification stage, we filter the quality based on four criteria outlined in the main text. For NLI filtering, we use mDeberta as our filtering model\footnote{The NLI model is available at \url{https://huggingface.co/MoritzLaurer/mDeBERTa-v3-base-xnli-multilingual-nli-2mil7}}, and filter out only samples predicted as "Contradiction" (approximately 15\%).

At the query level, we randomly select 16 ShareGPT samples for each instruction and perform Response Rejection Sampling with K=8. For instruction following verification, we adhere to the two standards mentioned in the text. Finally, for query quality verification, we filter for consistency using a threshold of 8.

\section{Case Study of Data Combination}\label{app:contamination_case_study}

\begin{figure*}[h]
    \centering
    \small    \includegraphics[width=0.98\linewidth]{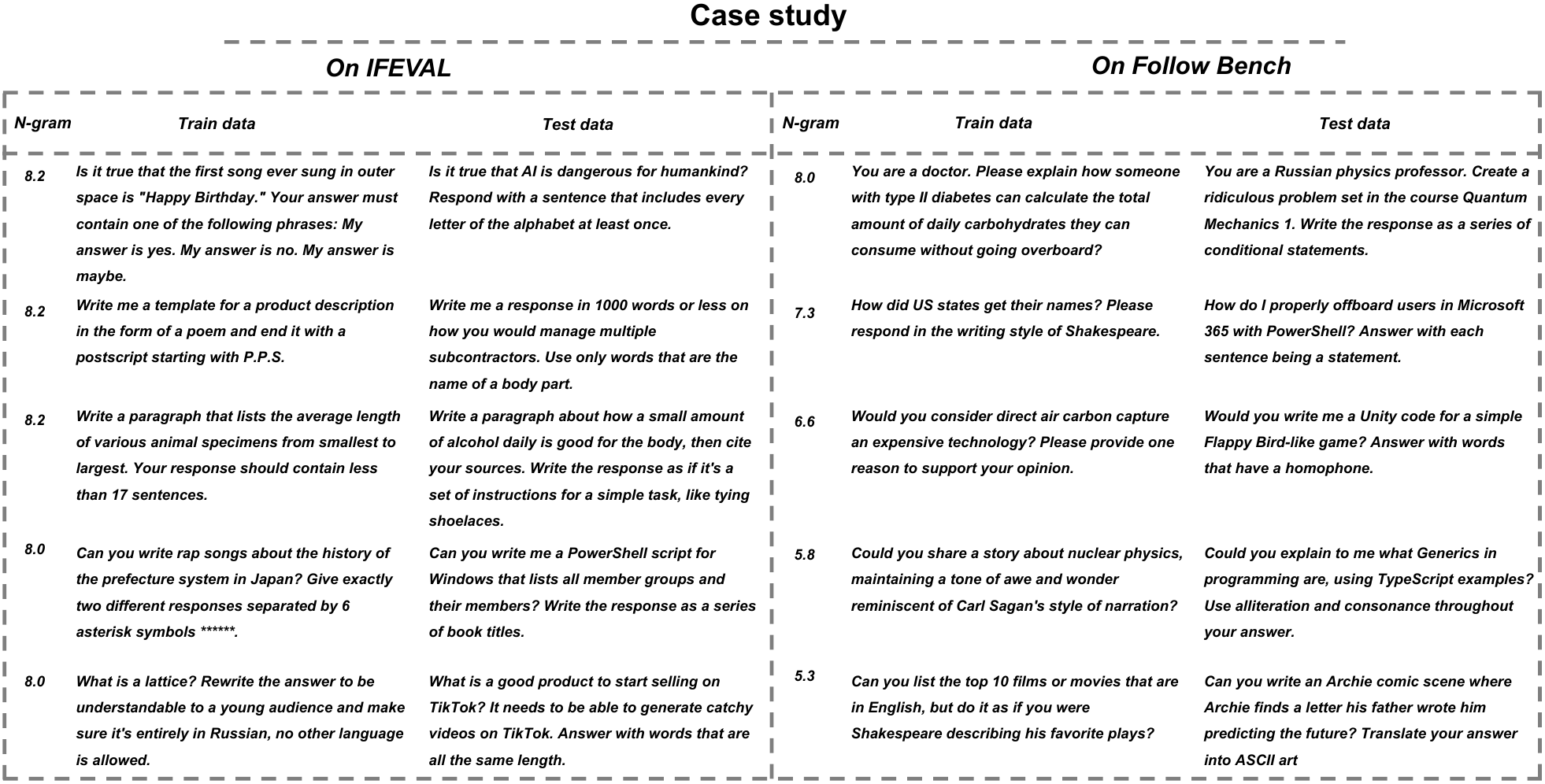}
    \caption{Case Study of data combination on IFEval and Followbench}\vspace{-12pt}
    \label{fig:case study}
\end{figure*}

We used n-gram 13 to evaluate the overlap between each test sample and the SFT training samples. It is unnecessary to evaluate DPO data since the inputs for DPO data are derived from SFT data. In \Cref{tab:contamination}, all our data combination metrics (both model-based and rule-based evaluation) are lower than those of ShareGPT, confirming that our method has no data combination with the test set. We also present the top 5 training-test sample overlaps in n-gram for both IF Eval and Followbench in \Cref{fig:case study}.

\section{Prompt Templates}
For the Self-Instruct stage, we use the following prompt template for instructions' rejection sampling:

\begin{tcolorbox}[
colback=white!10!white,
colframe=black!75!black,
title=Prompt Template of Self-Instruct Stage,
breakable]
You are an expert for writing instructions. Please provide \textbf{\{K\}} different instructions that meet the following requirements:\\
- Instructions are about the format but not style of a response\\
- Whether instructions are followed can be easily evaluate by a Python function\\
Here are some examples of instructions we need:\\
\textbf{\{Seed Instructions\}}\\
Do not generate instructions about writing style, using metaphor, or translation. Here are some examples of instructions we do not need:\\
- Incorporate a famous historical quote seamlessly into your answer\\
- Translate your answer into Pig Latin\\
- Use only words that are also a type of food\\
- Respond with a metaphor in every sentence\\
- Write the response as if you are a character from a Shakespearean play\\
Please generate one instruction per line in your response and start each line with '- '.\\
\end{tcolorbox}

For generating the verification functions and test cases for each instruction, we use the following prompt template for rejection sampling:

\begin{tcolorbox}[
colback=white!10!white,
colframe=black!75!black,
title=Prompt Template for Generating Verification Functions and Cases,
breakable]
You are an expert for writing evaluation functions in Python to evaluate whether a response strictly follows an instruction.\\
Here is the instruction: \textbf{\{instruction\}}\\
Please write a Python function named `evaluate` to evaluate whether an input string `response` follows this instruction. If it follows, simply return True, otherwise return False.\\
Please respond with a single JSON that includes the evaluation function in the key `func`, and a list of three test cases in the key `cases`, which includes an input in the key `input` and an expected output in the key `output` (True or False).\\
Here is an example of output JSON format: \\
\{\\
    "func": "JSON Str“, \\
    "cases": [
        \{ "input": "str", "output": "True" \},
        \{ "input": "str", "output": "False" \}
    ]\\
\}
\end{tcolorbox}

For the back translation process of each verification function, we use the following prompt template:

\begin{tcolorbox}[
colback=white!10!white,
colframe=black!75!black,
title=Prompt Template for Back Translation,
breakable]
You are an expert in converting Python eval function code into the corresponding instruction text. I will provide the eval function code. Please strictly follow the code to convert it into the corresponding instruction text. \\
Here's an example: \\
 \textbf{\{Example func\}} \\
 \textbf{\{Example cases\}}

Please convert the following eval function into instructions stored in a list: 
\\ 
 \textbf{\{funcs\}}
\end{tcolorbox}

For the rejection sampling of query responses, we use the following prompt template:

\begin{tcolorbox}[
colback=white!10!white,
colframe=black!75!black,
title=Prompt Template for Response Generation,
breakable]

Please answer the query strictly following the instruction. \\
Instruction: \{instruction\} \\
Query: \{query\}

\end{tcolorbox}

Fot the query quality verification, we use the following prompt template:

\begin{tcolorbox}[
colback=white!10!white,
colframe=black!75!black,
title=Prompt Template for Response Generation,
breakable]

You are an expert that is good at judging whether a response is following the instruction and query. \\
Instruction:  \textbf{\{instruction\}} \\
Query:  \textbf{\{query\}}  \\
Response:  \textbf{\{response\}}  \\
Please notice that the response may not be helpful as it needs to strictly follow the requirements in the Instruction. \\
You need to judge whether the response answers the query. Please first provide a detailed analysis and then give a score ranking from 0 to 10 at the last line. \\
Scoring 0 means the response is totally unrelated to the query, while scoring 10 means the response is helpful and highly related to the query. \\
Please only provide a score in the format `Score: {{score}}` without any other contents at the last line.
\end{tcolorbox}

\section{Case Study of \modelname}
In \Cref{tab:case-study}, we illustrates the data format of our \modelname, including the query, response (verification funcs Acc>0.8) and verification function.

\begin{table*}[h]
\centering
\small
\renewcommand{\arraystretch}{1.3}

\begin{tabular}{|m{3.5cm}|m{3.5cm}|m{5.2cm}|}
\hline
\textbf{Query} &\centering\textbf{Response}  &\centering\arraybackslash \textbf{Verification Funcs} \\
\hline
\centering Keep your answer to under 50 characters total. what is autoarima in python.  &\centering AutoARIMA automates ARIMA model selection. & \begin{lstlisting}[style=python]
def evaluate(response: str) -> bool:
    return len(response) <= 50
\end{lstlisting}
\\
\hline
\centering Refrain from using any words that contain 'S'. i need a name for my Dungeons and Dragons world.  &\centering EternaRealm & \begin{lstlisting}[style=python]
def evaluate(response):
    forbidden = 'sS'
    for char in response:
        if char in forbidden:
            return False
    return True
\end{lstlisting}
\\
\hline

\centering Keep your response under twenty words without sacrificing clarity. Let's play a game shall we?.  &\centering ure, let's play a game! What game do you have in mind? Please keep instructions simple and clear. & \begin{lstlisting}[style=python]
def evaluate(response: str) -> bool:
    return len(response.split()) <= 20 and len(response) > 0
\end{lstlisting}
\\
\hline
\centering Compose your answer using exactly 20 words. Diffrent Normal Forms.  &\centering Normal forms in databases: 1NF ensures atomic columns, 2NF eliminates non-key dependencies, 3NF removes transitive dependencies, BCNF enforces determinant restriction. & \begin{lstlisting}[style=python]
def evaluate(response: str) -> bool:
    return len(response.split()) == 20
\end{lstlisting}
\\

\hline

\centering Word your response in a five-line verse with a strict AABBA rhyme. 1.Write short notes on Decision trees..  &\centering Decision trees, so clear and bright, Branch out to split data's might, With nodes of questions, true or false, They sort through cases, young or old, like a versatile horse. From root to leaves, paths decide their course. & \begin{lstlisting}[style=python]
def evaluate(response: str) -> bool:
    lines = response.split('\n')
    if len(lines) != 5:
        return False
    rhymes = [line[-1] for line in lines]
    rhyme_pattern = rhymes[0] == rhymes[1] == rhymes[2] != rhymes[3] == rhymes[4]
    return rhyme_pattern
\end{lstlisting}
\\

\hline
% \centering\arraybackslash "def evaluate(response: str) -> bool:\n    return len(response) <= 50"

\end{tabular}

\caption{Examples of AutoIF's data formats.}
\label{tab:case-study}
\end{table*}

\section{Baselines \& Datasets\label{app:baselines}}

We give introductions to the LLM baselines for our instruction following.

\paragraph{LLaMA3}~\citep{llama3}, developed by MetaAI, is the latest iteration of the LLaMA series, featuring significant upgrades. Compared to LLaMA2, LLaMA3 expands its training dataset, context length, and vocabulary, resulting in improved performance across various tasks. Enhancements in contextual understanding and language generation further distinguish LLaMA3.

\paragraph{Qwen2}~\citep{bai2023qwen}, developed by Alibaba, includes five sizes: Qwen2-0.5B, Qwen2-1.5B, Qwen2-7B, Qwen2-57B-A14B, and Qwen2-72B. Trained on high-quality data in Chinese, English, and 27 other languages, Qwen2 excels in multilingual capabilities and shows strong performance in coding and mathematics. Additionally, it supports extended context lengths of up to 128K tokens (Qwen2-72B-Instruct), making it ideal for long texts and complex tasks.

\paragraph{Mistral-7B}~\citep{jiang2023mistral}, released by Mistral AI in September 2023, leverages grouped query attention (GQA) combined with sliding window attention (SWA) to efficiently process sequences of any length, enhance inference speed, and improve throughput. It outperforms many 13B models across various tasks.

\paragraph{Mixtral-8$\times$7B}~\citep{jiang2024mixtral} developed by Mistral AI, is the first open-source MOE large model. It is a sparse mixture of experts network and, like Mistral 7B, employs the GQA mechanism. With a smaller parameter count compared to LLaMA2-70B and GPT-3.5, it outperforms them across numerous tasks.

\paragraph{GPT Series} GPT-3.5~\citep{ChatGPT} and GPT-4~\citep{achiam2023gpt}, developed by OpenAI, are advanced models in the GPT series that use a three-stage reinforcement learning with human feedback (RLHF) algorithm. This enhances their instruction-following capabilities and minimizes harmful content generation. GPT-3.5 excels in text completion, translation, and summarization. Building on these strengths, GPT-4 further refines the RLHF algorithm, enhancing performance on complex instructions and making it suitable for applications ranging from academic research to industrial use.

%序号 ， train data / test data ， ngram similiary （IFEVAl case study （10条））
%序号 ， train data / test data ， ngram similiary （follow case study （10条））

In addition to the two Instruction-Following benchmarks introduced in the main text, we also provide a detailed overview of datasets covered in the experiments

\textbf{ShareGPT} refers to the multi-turn chatting histories used by Vicuna~\cite{vicuna2023}. ShareGPT includes 86K human queries and responses from ChatGPT and other chatbots. We randomly select 2w samples to train LLaMA3-8B and Qwen2-7B to obtain our baseline models:\textbf{ LLaMA3-8B (ShareGPT)} and \textbf{Qwen2-7B (ShareGPT)}.\footnote{Follow the set up of~\citeauthor{dong2023abilities}, we use the version from \url{https://huggingface.co/datasets/anon8231489123/ShareGPT_Vicuna_unfiltered} cleaned raw dataset, and follow Vicuna preprocess.}.

\textbf{GSM8K}~\citep{cobbe2021training} is a mathematical dataset designed to evaluate the mathematical problem-solving abilities of language models. It consists of 8,000 diverse grade school-level math word problems, which require understanding and manipulating mathematical concepts to arrive at a correct solution. It comprises high-quality grade school math problems, with 7,473 training samples and 1,319 testing samples.

\textbf{HumanEval}~\citep{chen2021evaluating} includes 164 unique programming challenges, each paired with approximately 9.6 test cases on average. To provide a more comprehensive evaluation of the functional accuracy of code generated by large language models, HumanEval+ substantially increases the number of test cases to an average of 774.8 per problem. In this paper, we report the Pass@1 result when applying greedy decoding.

\textbf{MMLU}~\citep{hendrycks2021measuring} is a benchmark designed to assess pretraining knowledge in models using zero-shot and few-shot evaluations. It includes 57 subjects across STEM, humanities, social sciences, and more, with difficulty levels ranging from elementary to advanced professional. MMLU tests both world knowledge and problem-solving skills, covering traditional disciplines like mathematics and history, as well as specialized areas such as law and ethics.

\textbf{C-Eval}~\citep{huang2023ceval} consists of multiple-choice questions categorized into four difficulty levels: middle school, high school, college, and professional. The questions cover 52 varied disciplines, including humanities, science, and engineering. Additionally, there is C-Eval Hard, a subset of particularly challenging topics within C-Eval that demand advanced reasoning skills. We perform an in-depth evaluation of leading language models on C-Eval, testing both English and Chinese-focused models.